\documentclass[twoside]{article}

\usepackage[accepted]{aistats2024}
\usepackage{hyperref}
\usepackage[round]{natbib}
\usepackage[round]{natbib}

\usepackage{amsfonts}       % blackboard math symbols
\usepackage{xcolor}         % colors

\usepackage{booktabs}

\usepackage{amsmath,amsfonts,amssymb,amsthm} % math environment
\usepackage{graphicx, caption, subcaption}
\usepackage{multirow}
\usepackage[normalem]{ulem}
\usepackage{tablefootnote}

\usepackage{soul}

\theoremstyle{plain}

\theoremstyle{definition}

\theoremstyle{remark}

\newcommand{\func}[1]{\mathtt{#1}}

\renewcommand{\cite}[1]{\citep{#1}}

\newcommand{\bx}{\mathbf{x}} % node feature vector 
\newcommand{\bW}{\mathbf{W}} % weight matrix 
\newcommand{\bZ}{\mathbf{Z}} % PE matrix 
\newcommand{\bz}{\mathbf{z}} % PE vector 

\def\gD{{\mathcal{D}}}

\def\gT{{\mathcal{T}}}

\def\sR{{\mathbb{R}}}

\def\bw{{\mathbf{w}}}
\def\by{{\mathbf{y}}}
\def\bY{{\mathbf{Y}}}
\def\bx{{\mathbf{x}}}

\def\bp{{\mathbf{p}}}

\begin{document}



\twocolumn[

\aistatstitle{Multi-resolution Time-Series Transformer for Long-term Forecasting}

\aistatsauthor{Yitian Zhang$^{1}$\textsuperscript{\dag} \And Liheng Ma$^{1}$\textsuperscript{\dag} \And  Soumyasundar Pal$^2$\And Yingxue Zhang$^2$ \And Mark Coates$^{1}$}
% \aistatsaddress{  $^1$ McGill University, Montreal, Canada \\ $^2$ Huawei Noah’s Ark Lab, Montreal, Canada \\
% $^3$ Mila - Quebec AI Institute \\
% $^4$International Laboratory on
% Learning Systems (ILLS)} 
\aistatsaddress{  $^1$McGill University, Mila, and ILLS \quad \quad $^2$Huawei Noah's Ark Lab} ]

\begin{abstract}
% Motivated by the success of attention mechanisms in the modeling of sequential data in natural language processing and computer vision, there have been extensive research efforts for using transformer architectures for time-series forecasting in recent years. 
%Recent work on transformers for time-series forecasting suggests that the incorporation of local semantic information by segmentation of time-series and use of the resulting patches as tokens facilitates capturing complex temporal patterns present in time-series.
% real-world datasets. 
%The performance of transformers for time-series forecasting has improved significantly through the incorporation of local semantic information. 
The performance of transformers for time-series forecasting has improved significantly. Recent architectures learn complex temporal patterns by segmenting a time-series into patches and using the patches as tokens. The patch size controls the ability of transformers to learn the temporal patterns at different frequencies:
shorter patches are effective for learning localized, high-frequency patterns, whereas mining long-term seasonalities and trends requires longer patches.
Inspired by this observation, we propose a novel framework, \textbf{M}ulti-resolution \textbf{T}ime-\textbf{S}eries \textbf{T}ransformer (MTST), which consists of a multi-branch architecture for simultaneous modeling of diverse temporal patterns at different resolutions.
In contrast to many existing time-series transformers, we employ relative positional encoding, which is better suited for extracting periodic components at different scales. 
Extensive experiments on several real-world datasets demonstrate the effectiveness of MTST in comparison to \emph{state-of-the-art} forecasting techniques. 
% \YX{Lack a transition sentence from multi-resolution idea to relative position encoding idea. Seem two completely isolated ideas.}
% \YX{Summarize the quantitative results comparing to other SOTA method.}
% Transformer-based models have demonstrated outstanding performance on long-term time-series forecasting with the patching scheme, which treats subsequences of each variable as tokens of self-attentions (SAs). 
% In this research, we reveal that the choices of patch size will affect the ability of transformers to capture the periodic patterns of different frequencies in time series: SAs with smaller patches are more conscious to high-frequency patterns but prone to omit the low-frequency patterns, while the larger patches enable the model to capture longer-term patterns and trends, including low-frequency patterns.
% Thus, we propose a novel architecture, Multi-resolution Time-Series Transformers (MTSTs), to capture patterns of different frequencies. MTST introduces a multi-resolution learning scheme by simply adjusting the patch sizes. 
% The proposed multi-branch multi-resolution self-attention module results in a more flexible and powerful way to capture temporal information. Extensive experimental results on eight real-world datasets show the effectiveness of MTSTs against previous state-of-the-art baselines.
\end{abstract}

\section{INTRODUCTION}
\label{section:intro}
Time-series forecasting has ubiquitous applications in various domains including but not limited to quantitative finance, weather prediction, and electricity management.
% the development of TSTs
% After the surge of transformers in diverse fields, e.g., natural language processing
% ~\citep{vaswani2017AttentionAllYou, dai2019TransformerXLAttentiveLanguage, devlin2019BERTPretrainingDeep, radford2018ImprovingLanguageUnderstanding}, computer vision~\citep{dosovitskiy2020ImageWorth16x16, liu2021SwinTransformerHierarchical,wang2021PyramidVisionTransformer,fan2021MultiscaleVisionTransformers},
% transformer architectures are introduced in multivariate time-series forecasting~\citep{zhou2021InformerEfficientTransformer, wu2021AutoformerDecompositionTransformers, zhou2022FEDformerFrequencyEnhanced}, viewing each timestamp as a token.
Building on the success of transformers in diverse fields, %including natural language processing, computer vision and protein modeling,
transformer architectures were recently introduced in multivariate time-series forecasting, viewing each timestamp as a token~\citep{zhou2021InformerEfficientTransformer, wu2021AutoformerDecompositionTransformers, zhou2022FEDformerFrequencyEnhanced}.
However, the effectiveness of time-series transformers (TSTs) is disputed: most TSTs are not sensitive to the temporal order of tokens and can even be outperformed by simple linear models~\citep{zeng2023AreTransformersEffective}.
% on their insensitivity to time-series order and inability to model temporal patterns.
% Furthermore, 
Recently, the work of \citet{zhang2023CrossformerTransformerUtilizing} and \citet{nie2023TimeSeriesWorth} revealed that 
timestamp-level tokenization prevents attention mechanisms from effectively capturing temporal patterns.
In contrast,
patch-level tokenization (where a \emph{patch} is a window of timesteps) allows attention mechanisms to model temporal patterns within each patch and learn the relationships between patches.

\begin{figure*}[htbp]
    \centering
\includegraphics[width=0.975\textwidth]{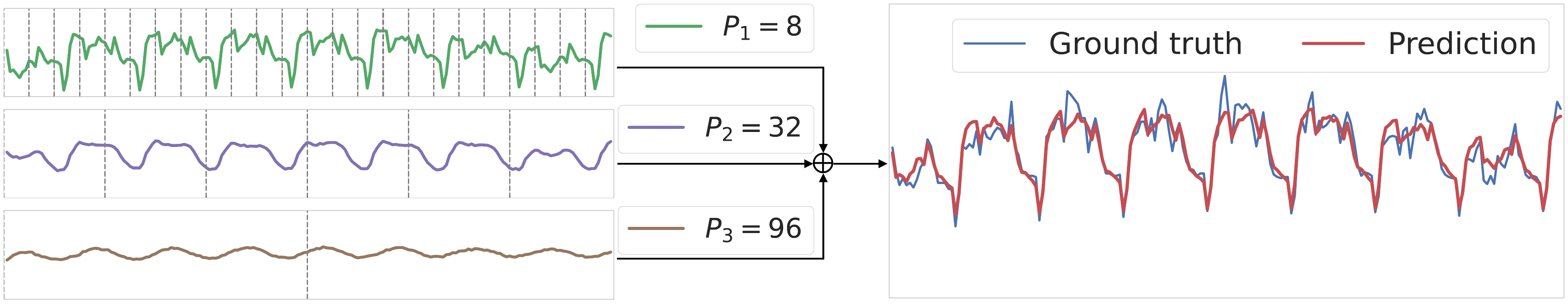}
    \caption{An example from \textit{Electricity} dataset: MTST learns multi-scale temporal patterns in different branches, where $P_i$ stands for the patch size in $i$-th branch.}
    % \vspace{-1em}
    \label{fig:multiscale}
\end{figure*}
% problematic when different behaviors are exhibited in different variates (e.g., asynchronicity and low-correlation).
% Accordingly, 
% inspired by vision transformers~\citep{dosovitskiy2020ImageWorth16x16}, 
% they propose to construct tokens by patching the time-series into sub-series for each channel independently, triggering the come-from-behind victory of TSTs.

% \def\starfootnote{*}\footnotetext{Equal contribution. Work done as interns at Huawei Noah’s Ark Lab, Montreal Research Center.}

% Footnote explaining the cross symbol
% \noindent
% \textsuperscript{†}These authors contributed equally to this work.
% \footnotetext{\textsuperscript{†}These authors contributed equally to this work.}

% Manually add the unnumbered footnote using \dag for the cross symbol
\begingroup
\renewcommand\thefootnote{\dag} 
% \footnotetext{Equal contribution. Work done as interns at Huawei
% Noah’s Ark Lab, Montreal Research Center.}
\footnotetext{Equal contribution, the work was partially done when YZ and LM were interns at Huawei Noah's Ark Lab.}
\endgroup

Despite the promising performance demonstrated by patch-based TSTs,
these methods fail to explicitly incorporate multi-scale analysis, which has proved effective in many time-series modeling domains.
Therefore, in this work, we endow patch-based TSTs with the ability to learn multi-scale features with attention mechanisms via multi-resolution representations,
and propose a novel architecture, \textbf{M}ulti-resolution \textbf{T}ime-\textbf{S}eries \textbf{T}ransformer (MTST)~\footnote{Our code is publicly available at \href{https://github.com/networkslab/MTST}{https://github.com\\/networkslab/MTST}.}.
Unlike previous works that rely on subsampling, MTST constructs a multi-resolution representation by simply adjusting the patch-level tokenization of a time-series: 
a large number of small-sized patches leads to high-resolution feature maps; 
a small number of large-sized patches results in low-resolution feature maps.
By constructing multiple sets of tokens with different patch-sizes, 
each MTST layer can model the temporal patterns of different frequencies simultaneously with multi-branch self-attentions.
As shown in an example from the Electricity dataset (Figure~\ref{fig:multiscale}), the role of the branch with larger-size patches is mostly to capture the lower-frequency and coarser temporal patterns; the branch with smaller-size patches contributes to modeling the higher-frequency and finer temporal patterns.
By processing the signals with a multi-resolution multi-branch architecture, MTST can model complex temporal signals that contain multiple seasonalities.
Furthermore, in order to overcome the weak sensitivity to the ordering of time-series that is exhibited by many TSTs,
instead of employing learned/fixed absolute positional encoding,
we utilize \emph{relative} positional encoding, which is naturally aligned with capturing periodic temporal patterns.

To provide support for our motivation and hypotheses, we conduct extensive experiments on multiple time-series forecasting benchmarks.
Our proposed MTST demonstrates \textit{state-of-the-art} performance in comparison with diverse forecasting methods, reaching the lowest mean squared error on 28 out of 28 test settings.
Comprehensive ablation studies and analysis experiments justify the effectiveness of proposed designs and the consequent advantage over previous techniques.

\section{PROBLEM STATEMENT}
\label{sec:Problem}
We focus on long-term forecasting of multivariate time-series from historical data.
Let $\bx_t{=}[x_{t,1}, x_{t,2}, \cdots, x_{t,M}]^\intercal \in \sR^{M \times 1}$ be a multivariate signal, where $x_{t,m}$ denotes $m$-th variate at time $t$, for $1 {\leqslant} m {\leqslant} M$.
The goal is to learn a model which can forecast the future $T$ timestamps from the recent history of $L$ timestamps.
Here, $L$ and $T$ are termed the look-back window and prediction horizon, respectively.
In other words, for any arbitrary time offset $t_0$, the model processes $\bx_{t_0+1:t_0+L}$ as input and provides an estimate of $\bx_{t_0+L+1:t_0+L+T}$ as its output.
The estimate of $\bx_t$ is denoted by $\widehat{\bx}_t$.
We drop the time offset $t_0$ in all subsequent discussions for simplicity. %conciseness. 

A training dataset $\gD_{trn}$ is assumed to be available for learning the model parameters. Usually, the time-series is spliced to construct the training set. The $k$-th example in the training set is denoted by $(\bx_{1:L}^{(k)}, \bx_{L+1:L+T}^{(k)})$.
While $\bx_{1:L+T}$
is known in the training set, 
$\bx_{L+1:L+T}$ is unknown and must be estimated in the test set. 

The forecasting performance of the model is assessed by computing the mean squared error (MSE) and the mean absolute error (MAE) between the prediction and the ground truth on the test set $\gD_{test}$. These metrics are defined as:
\begin{align}
\text{MSE} &= \frac{1}{MT\lvert\gD_{test} \rvert}\sum_{k \in \gD_{test}} \sum_{t=L+1}^{L+T} \lvert \lvert \bx_t^{(k)} - \widehat{\bx}_t^{(k)} \rvert \rvert_2^2\,,\label{eq:mse}\\
\text{MAE} &= \frac{1}{MT\lvert\gD_{test} \rvert}\sum_{k \in \gD_{test}} \sum_{t=L+1}^{L+T} \lvert \lvert \bx_t^{(k)} - \widehat{\bx}_t^{(k)} \rvert \rvert_1\,.\label{eq:mae}
% \text{ }\nonumber
\end{align}

% Recent work has suggested that for many multivariate time-series of interest, 
% building a univariate forecasting model using a transformer with patches as tokens achieves \emph{state-of-the-art} performance~\citep{nie2023TimeSeriesWorth}. 
% In other words, each $\bx_{1:L,m}$ is processed independently to generate the output $\widehat{\bx}_{L+1:L+T,m}$. The univariate forecasts are subsequently combined to form the multivariate forecast.
% This property of a model is termed {\em channel-independence}. We follow the same approach and omit the variate index $m$ in order to simplify the notation. The extension of our work to incorporate channel dependencies is straightforward, but we do not explore the performance of the extension in this work. 

\begin{figure*}[htbp]
\centering
\includegraphics[width=0.975\textwidth]{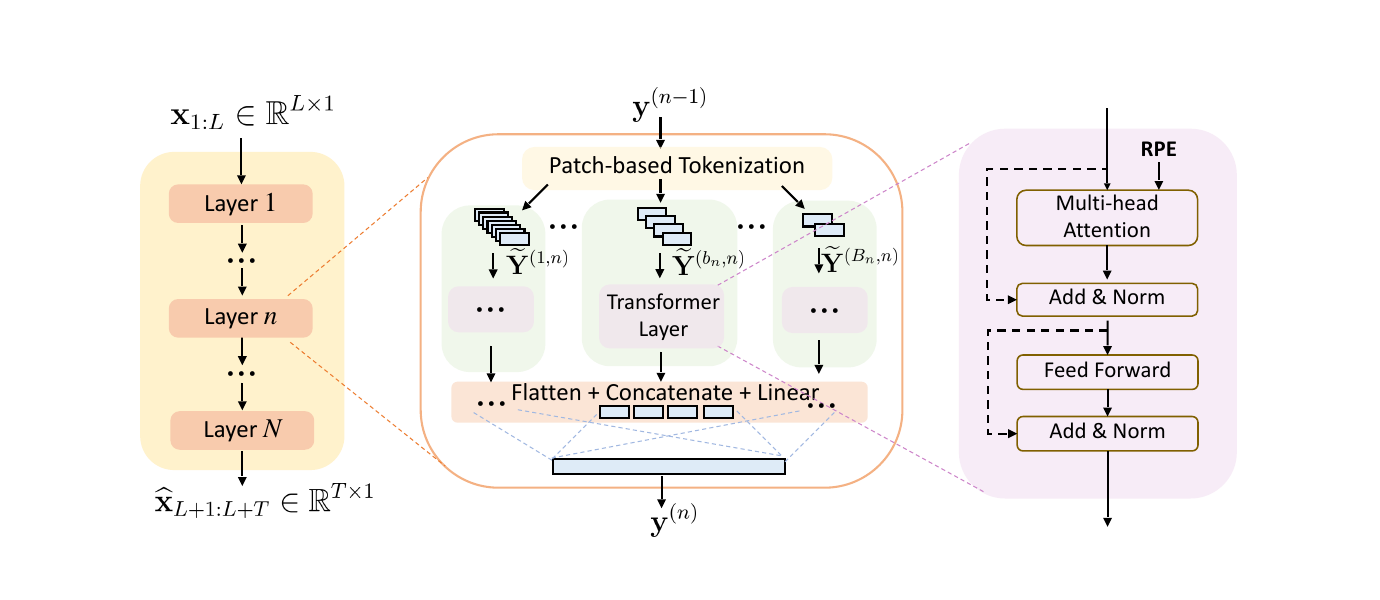}
\caption{Multi-resolution Time-Series Transformer (MTST) Architecture.}
\label{fig:MTST}
\end{figure*}

\begin{figure*}[ht!]
\centering
    \begin{minipage}[t]{0.56\textwidth}
     % \begin{minipage}{0.56\textwidth}
      % \includegraphics[width=\textwidth]{Figure/RPE/attn_83.pdf}
      % \vspace{8.5em}
      \includegraphics[scale=0.18]{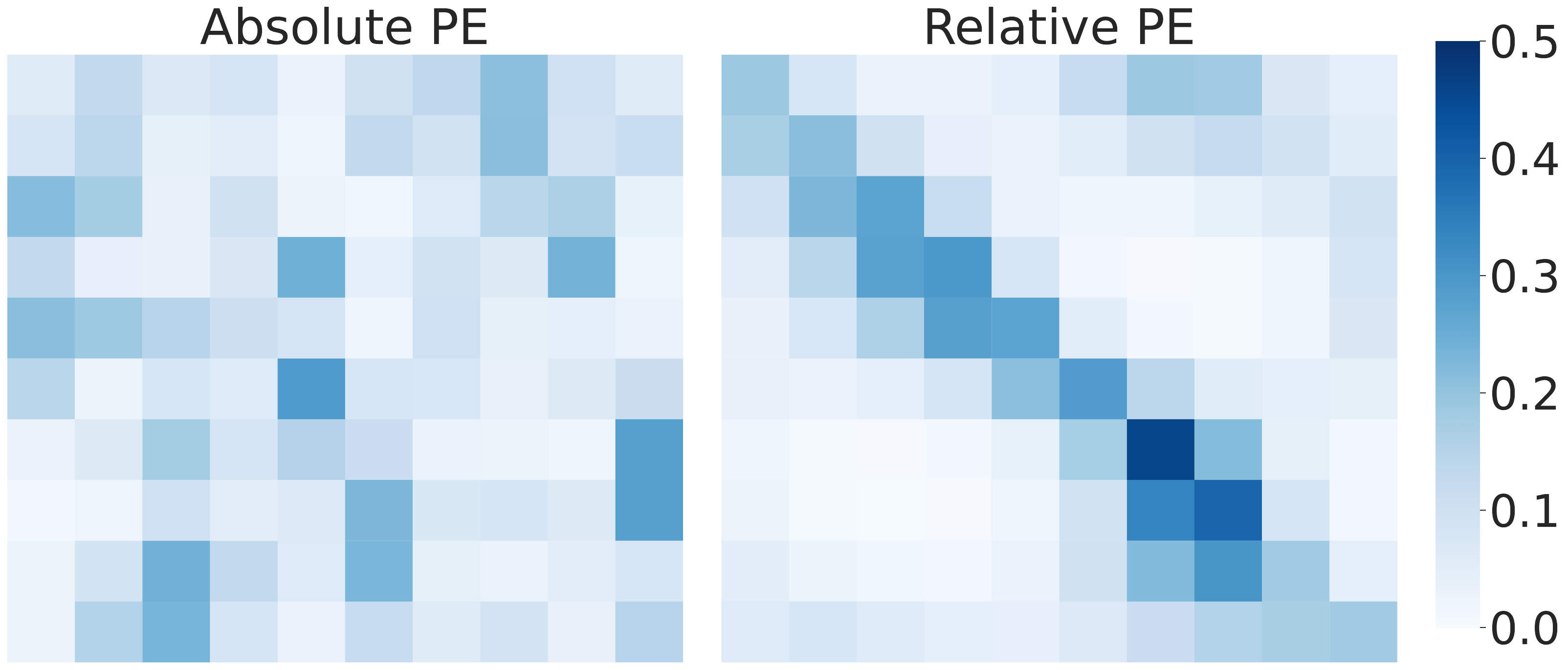}
      \subcaption{}
    \end{minipage} \hfill
    \begin{minipage}[t]{0.43\textwidth}
      \includegraphics[width=1.0\textwidth]{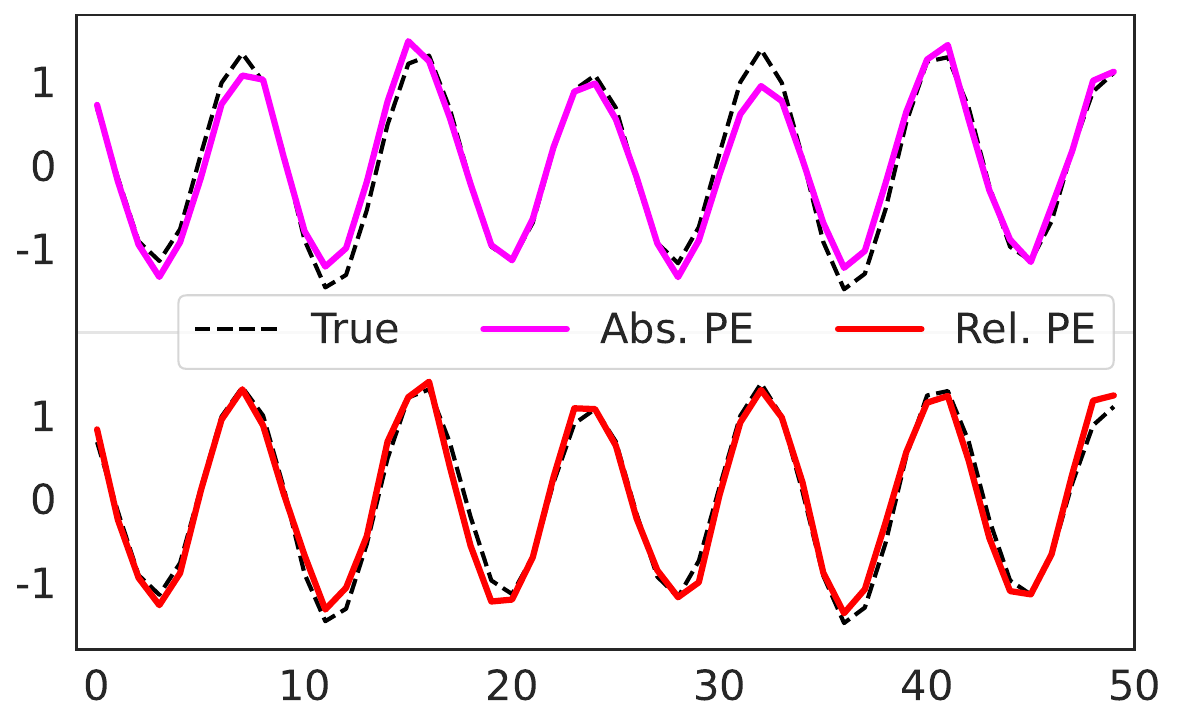}
      % \vspace{-0.5cm}
      \subcaption{}
    \end{minipage} 
        % \vspace{-1em}
      \caption{(a) Heatmaps of attention scores for absolute and relative positional encoding, (b) obtained forecasts using a transformer with absolute and relative positional encoding.}
    \label{fig:RPE}
\end{figure*}

\section{METHODOLOGY}
\label{sec:method}
The proposed MTST consists of $N$ layers, as shown in Fig.~\ref{fig:MTST}. 
Each MTST layer has multiple branches, with $B_n$ branches at the $n$-th layer. 
% \YX{Please double check all the notations to make sure the notation in text is consistent with the notation in the figure. For example, you use $N$ for the total number of layes, but use $K$ in the image.}
The multi-branch architecture allows us to learn representations of the time-series at different scales simultaneously.
Each branch contains a tokenizer (with different patch-size), which converts the input representation into patches.
These patches are further processed as tokens by self-attention (SA) with relative positional encoding (RPE). 
The branch outputs are fused together to form a single embedding to be fed to the next MTST layer.
% The extension of our work to incorporate channel dependencies is straightforward, but we do not explore the performance of the extension in this work. 
We delve into a detailed discussion of each of these components in this section.
% We discuss each of these components in detail next.

Following the protocol in the previous works~\cite{zeng2023AreTransformersEffective, nie2023TimeSeriesWorth}, MTST processes each $\bx_{1:L,m}$ independently to generate the output $\widehat{\bx}_{L+1:L+T,m}$, and subsequently combine them to form a multivariate forecast.
% The univariate forecasts are subsequently combined to form the multivariate forecast.
This technique is termed \textit{channel-independence} and we omit the variate index $m$ in order to simplify the notation in the following sections. 
However, MTST is not limited by channel-independence and is readily extended to other patch-based TSTs that model dependencies among variates, such as the Crossformer~\cite{zhang2023CrossformerTransformerUtilizing}.

\subsection{Branch specific tokenization}
\label{subsec:token}
Temporal patches consisting of multiple successive timestamps are essential for learning effective representations for forecasting~\citep{nie2023TimeSeriesWorth}. Let $\by^{(n-1)} {\in} \sR^{d_{n-1} {\times} 1}$ be the $d_{n{-}1}$-dimensional output representation of a univariate time-series at the $(n{-}1)$-th layer of MTST.
We have $d_0 {=} L$  and $\by^{(0)}{=}[x_1, \cdots, x_L]^\intercal$. Similarly, at the last layer, we have $d_N{=}T$ and $\by^{(N)}{=}[\widehat{x}_{L+1}, \cdots, \widehat{x}_{L+T}]^\intercal$.

Denote by $P_{b_n}$ the patch size and by $S_{b_n}$ the stride (length of the non-overlapping region between two successive patches) at the $b_n$-th branch. Then the tokenizer $\gT_{b_n}: \sR^{d_{n} {\times} 1} {\rightarrow} \sR^{J_{b_n} {\times} P_{b_n}}$ converts $\by^{(n-1)}$ into $J_{b_n} {=} \lceil (d_{n-1}{-}P_{b_n})/S_{b_n} \rceil + 1$ overlapping patches in a sliding-window fashion.
Specifically,  we compute $\widetilde{\bY}^{(b_n, n)} {=} [\widetilde{\by}_1^{(b_n, n)}, \cdots, \widetilde{\by}_{J_{b_n}}^{(b_n, n)}]^\intercal {=}  \gT_{b_n}(\by^{(n-1)})$ as follows:
\begin{align}
\widetilde{\by}_{j}^{(b_n, n)}{=}\by^{(n-1)}_{[(j-1)S_{b_n}+1:(j-1)S_{b_n}+P_{b_n}]}\,,
% 1 {\leqslant} j {\leqslant} J_{b_n}
\label{eq:patch} 
\end{align}
for 
$j=1,\cdots, J_{b_n}$.
Here $\widetilde{\bY}^{(b_n, n)}$ denotes the output of the tokenizer and $\widetilde{\by}_{j}^{{(b_n, n)}^\intercal}$ is its $j$-th row (token).
In order to make sure that the number of tokens is an integer, we pad the last dimension of $\by^{(n-1)}$ at the end of the last token $(J_{b_n}{-}1)S_{b_n}{+}P_{b_n}{-}d_{n-1}$ times, if $(d_{n-1}{-}P_{b_n})$ is not divisible by $S_{b_n}$.

A high value of $J_{b_n}$ (equivalently, a low value of $P_{b_n}$) allows the $b_n$-th branch to focus on shorter patches, resulting in higher-resolution modeling of short-term temporal features. By contrast, longer patches facilitate learning of longer-term seasonalities and trends.

\subsection{Self-attention}
\label{subsec:transformer}
The self-attention in MTST is performed on the patch-level tokens in each branch independently. We omit the layer index $n$ and branch index $b_n$ temporarily to simplify notation.  $\func{Attn}: \sR^{J {\times} P} {\rightarrow} \sR^{J {\times} D}$ is applied to the tokens to capture the relationships between them.
Specifically, we compute $\widetilde{\bZ} {=} [\widetilde{\bz}_1, \cdots, \widetilde{\bz}_J]^\intercal {=} \func{Attn}(\widetilde{\bY})$ as follows:
\begin{align}
\alpha_{ij} &{=} \func{Softmax}_{j} \Big(\frac{(\bW_Q\widetilde{\by}_i)^T (\bW_K\widetilde{\by}_j)}{\sqrt{D}} + \bw_{pos}^T\bp_{ij} \Big)\,,\label{eq:attn_score}\\
\widetilde{\bz}_i &{=} \sum_j \alpha_{ij} \bW_V \widetilde{\by}_j\,.\label{eq:attn_output}
\end{align} 
Here, $\bp_{ij}{\in}\sR^{D_\text{pos} {\times}1}$ and $\alpha_{ij}{\in} [0,1]$ represent the relative positional encoding and the attention score between the $i$-th and $j$-th tokens, respectively; $\bW_\text{Q}, \bW_\text{K}, \bW_\text{V} {{\in}}\sR^{D {\times} P}$ and $\bw_\text{pos} {{\in}}\sR^{D_\text{pos} {\times} 1}$ are learnable weights.
As in~\citep{vaswani2017AttentionAllYou}, we set $D{=}P$ and use a multi-head variant of this attention mechanism with different sets of weight matrices in each head. 
Moreover, similar to other transformer layers, the output of the self-attention operation is processed with a particular combination of residual connections, normalizations, and a feed-forward network. 
These details are shown schematically in Figure~\ref{fig:MTST}, but are deferred to the Appendix (Section~\ref{subsec:layer_details}) to avoid notational clutter. 

\subsubsection{Relative positional encoding}

Instead of absolute PE, employed in most previous TSTs, we introduce {\em relative} PE (RPE), which encodes the distances between each pair of tokens.
Figure~\ref{fig:RPE} demonstrates that the use of relative PE
enables each token to identify similar tokens
on a synthetic dataset with strong periodic patterns
with considerably better accuracy.
This results in improved forecasting.

Our relative PE is defined as $\bp_{ij}{:=} \func{sign}(i-j) \func{PE}(|i-j|)$,
where $\func{sign}$ denotes the sign function and the function $\func{PE}: \mathbb{Z}^+{\rightarrow}\sR^{D_{pos}}$ is defined as
\begin{align}
\func{PE}_{2t}(i)&{:=} \sin(i/10000^{2t/D_{pos}})\,,\nonumber\\
\func{PE}_{2t{+}1}(i)&{:=} \cos(i/10000^{2t/D_{pos}})\,,\label{eq:pe_design}
\end{align}
for $t \in \{1, \ldots, D_{pos}/2\}$.

\subsection{Fusing representations from all branches}
\label{subsec:fuse}
At each layer, the token representations obtained from all branches are fused to form a single embedding. 
This allows sharing of information across scales, which helps in learning expressive representations of the time-series. 
The fusing is carried out by successive application of flattening, concatenation, and a  linear transformation.
First each $\widetilde{Z}^{(b_n, n)} {\in} \sR^{J_{b_n} {\times} P_{b_n}}$ is flattened to a row vector of length ${J_{b_n}{\cdot} P_{b_n}}$.
Next, the flattened vectors are concatenated to form $\bz^{(n)}=[\func{Flatten}(\widetilde{Z}^{(1, n)}), \cdots, \func{Flatten}(\widetilde{Z}^{(B_n, n)})]^T \in \sR^{\big(\sum_{b_n=1}^{B_n} J_{b_n}{\cdot}P_{b_n}\big) \times 1}$.
Finally a linear layer, with weight $\bW^{(n)} {\in} \sR^{d_{n} {\times} \big(\sum_{{b_n}=1}^{B_n} J_{b_n}{\cdot}P_{b_n}\big)}$ 
and bias $\mathbf{b}^{(n)} \in \sR^{d_{n}{\times}1}$, is used to obtain $\by^{(n)}$, i.e.,
\begin{align}
\by^{(n)} = \bW^{(n)} \bz^{(n)} + \mathbf{b}^{(n)} \,.\label{eq:linear_fuse}
\end{align}
We refer to the combination of operations as $\func{Fuse}$. The $n$-th layer of the MTST can be summarized as:
{\small
\begin{align}
\by^{(n)}{=} \func{Fuse}{\Big(}\func{Attn}{\big(}\gT_{1}{(}\by^{(n{-}1)}) {\big)}, {\ldots}, \func{Attn}{\big(}\gT_{B_n}(\by^{(n{-}1)}) {\big)} {\Big)}\,.
\end{align}}

\section{RELATED WORK}
\label{sec:rel}

% \mlh{Mark's comment:
% I don’t think you stress the novelty or the deficiencies of existing methods enough. I would suggest writing a sentence in italics at the end of each section that summarizes what the weaknesses are and how your approach fixes them.}

% \mlh{We potentially need to mention a bit on channel-independent if not citing them in methodology.}

\subsection{Long-horizon Time-Series Forecasting}
Multivariate time-series forecasting has been an active research area for decades.
Until recently, statistical modeling based algorithms were the most effective, but  many deep learning techniques have emerged and achieved impressive forecasting~\cite{oreshkin2021NbeatsNeuralNetwork, salinas2020DeeparProbabilisticForecasting,sen2019ThinkGloballyAct}.
However, convolutional and recurrent architectures fail to capture long-range dependency, leading to poor long-term forecasting. 
In order to address this, transformer-based models have been proposed.
Earlier versions treated each timestamp as a token and performed forecasting in a sequence-to-sequence manner~\cite{li2019EnhancingLocalityBreaking, zhou2021InformerEfficientTransformer}.
Several time-series transformers (TSTs), such as  Autoformer~\cite{wu2021AutoformerDecompositionTransformers} and FEDformer~\cite{zhou2022FEDformerFrequencyEnhanced}, incorporated inductive biases such as trends and seasonalities. 

\citet{zeng2023AreTransformersEffective} demonstrated that timestamp-level tokenization  prevents models from capturing temporal patterns. A simple linear model outperforms most timestamp-level TSTs.
Crossformer~\cite{zhang2023CrossformerTransformerUtilizing} and PatchTST~\cite{nie2023TimeSeriesWorth} address this by using patches (windows of multiple timesteps) as tokens, inspired by the patch-based transformer for images in~\cite{dosovitskiy2020ImageWorth16x16}. 

Although previous patch-based methods achieve excellent forecasting, they cannot disentangle multi-scale features.
% and rely on absolute PE, which is poorly aligned to the ubiquitous periodicities in time-series.
In contrast, our proposed MTST is designed to perform multi-resolution decomposition to naturally extract multiple periodicities.

\subsection{Multi-scale Feature Learning}
Several recent time-series forecasting techniques have incorporated multi-scale analysis.
NHiTS~\cite{challu2022NHiTSNeuralHierarchical} introduces multi-rate signal sampling and hierarchical interpolation to model features at multiple granularities. MICN~\cite{wang2023MicnMultiscaleLocal} incorporates convolution with different kernel sizes to learn multi-scale features.
TimesNet~\cite{wu2022TimesnetTemporal2dVariation} strives to capture the multi-periodicity in time-series by converting 1D sequences to a set of 2D tensors.
Among transformer-based models, 
Pyraformer~\cite{liu2021PyraformerLow-complexityPyramidal} forms a multi-resolution representation via pyramidal attention. Scaleformer~\cite{shabani2023Scaleformer} proposes a multi-resolution representation framework for existing timestamp-based TSTs via down/up-sampling.
However, migrating existing multiscale techniques to token-based TSTs is not trivial: the subsampling/downsampling techniques in previous work  result in sub-optimal representations for patches, because the methods are not cognizant of the chronological order at the timestamp level~\cite{marin2023TokenPoolingVision}.

Although these methods have tried to incorporate multi-resolution analysis, the attempts have proved ineffective. Pyraformer and Scaleformer’s forecasts are dramatically inferior to those of PatchTST, which is incapable of forming explicit multi-scale decompositions.

\subsection{Positional Encoding} 
To enable transformers to sense token positions, 
learned or fixed sinusoidal absolute positional encoding (PE) is usually injected into tokens before the encoders~\cite{vaswani2017AttentionAllYou}.
\citet{huang2019MusicTransformerGenerating} observe that relative PE~\cite{shaw2018SelfAttentionRelativePosition} can better model periodic patterns, which is crucial in time-series. In contrast to all previous TSTs that use absolute PE,
MTST employs a flexible relative PE~\cite{dai2019TransformerXLAttentiveLanguage}, whose design provides useful inductive bias for forecasting.

\section{EXPERIMENTS}
\label{sec:exp}

\subsection{Benchmarking MTST}

\subsubsection{Datasets}
We evaluate the performance of our proposed MTST on seven widely-used
public benchmark datasets, including Weather, Traffic, Electricity and four ETT datasets (ETTh1, ETTh2, ETTm1, ETTm2).
\textit{Weather} is a collection of 2020 weather data from 21 meteorological indicators, including air temperature and humidity, provided by the Max-Planck Institute for Biogeochemistry\footnote{https://www.bgc-jena.mpg.de/wetter}.
\textit{Traffic} is a dataset provided by Caltrans Performance Measurement System (PeMS), collecting hourly data of the road occupancy rates measured by different sensors on San Francisco Bay area freeways from California Department of Transportation\footnote{https://pems.dot.ca.gov}.
\textit{Electricity} contains hourly time-series of the electricity consumption of 321 customers from 2012 to 2014~\cite{trindade2015ElectricityLoadDiagrams, wu2021AutoformerDecompositionTransformers}\footnote{\citet{wu2021AutoformerDecompositionTransformers} selected 321 of 370 customers from the original dataset in~\citet{trindade2015ElectricityLoadDiagrams}. This version is widely used in the follow-up works.}.
\textit{ETT} datasets are composed of a series of sensor measurements, including load and oil temperature, collected from electricity transformers between 2016 and 2018,  provided by \citet{zhou2021InformerEfficientTransformer}. 
Following the standard pipelines, the datasets are split into training, validation, and test sets with the ratio of $6{:}2{:}2$ for four ETT datasets and $7{:}1{:}2$ for the remaining datasets.
Detailed statistics of the datasets are summarized in Table~\ref{tab:datasetStat}.

\begin{table}[ht!]
    \centering
    \scriptsize
    \setlength{\tabcolsep}{2pt}
    \begin{tabular}{c|ccccc}
    \toprule
Datasets & ETTh1/2 & ETTm1/2 & Traffic & Electricity & Weather \\ 
    \midrule
Variates & 7 & 7 & 862 & 321 & 21 \\
Timesteps & 17,420 & 69,680 & 17,544 & 26,304 & 52,696  \\
Granularity &1 hour & 15 min &1 hour &1 hour &10 min\\
    \bottomrule
    \end{tabular}
    \caption{The statistics of datasets used in the experiments}
    \label{tab:datasetStat}
    % \vspace{-0.5em}
\end{table}

\begin{figure}[ht!]
\centering
\hspace{-1em}
% \vspace{-1em}
\includegraphics[width=0.47\textwidth]{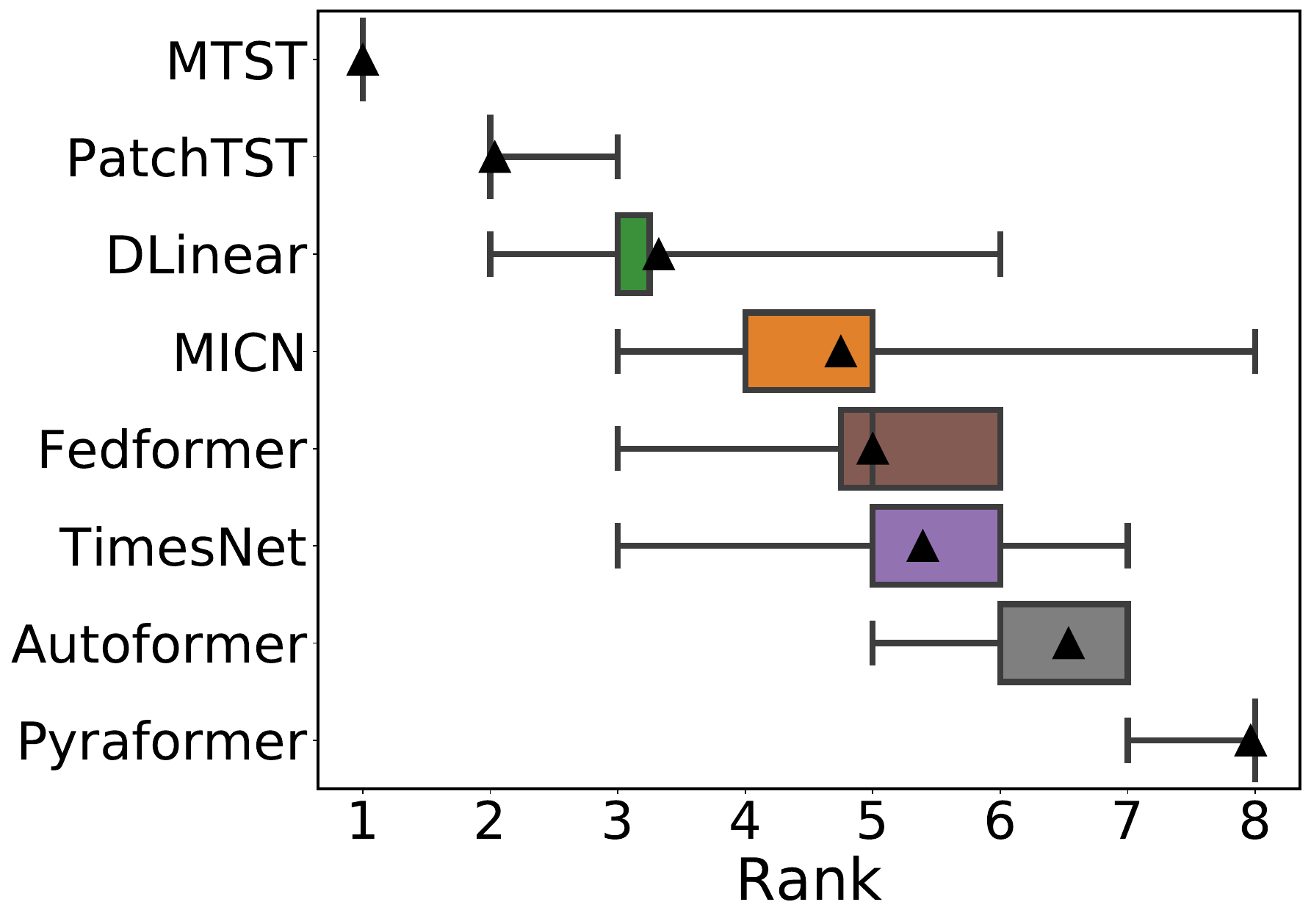}
\caption{Boxplot for ranks of the algorithms (based on their MSE) across seven datasets and four prediction horizons. The medians and means of the ranks are shown
by the vertical lines and the black triangles respectively;
whiskers extend to the minimum and maximum ranks.}
\label{fig:rank}
\end{figure}
\begin{table*}[ht!]
\centering
\scriptsize
\caption{
% \yitian{Multivariate long-term forecasting results with $T \in \{24, 36, 48, 60\}$ for ILI dataset and 
% $T \in \{96, 192, 336, 720\}$ for the others}.
Multivariate long-term forecasting results with $L=336$ and $T \in \{96, 192, 336, 720\}$.
\textbf{Bold} and \ul{underlined} denote the best and second-best results respectively.
$^*$ indicates statistically significant difference between the top-2 results.
% (see \S\ref{sec:main-res}).
}

\label{tab:L336}
\useunder{\uline}{\ul}{}
\setlength{\tabcolsep}{3pt}
\begin{tabular}{cc|ll|cc|cc|cc|cc|cc|cc|cc}
\toprule
\multicolumn{2}{c|}{Models}                             & \multicolumn{2}{c|}{MTST}       & \multicolumn{2}{c|}{PatchTST} & \multicolumn{2}{c|}{DLinear} & \multicolumn{2}{c|}{MICN}      & \multicolumn{2}{c|}{TimesNet}    & \multicolumn{2}{c|}{Fedformer}   & \multicolumn{2}{c|}{Autoformer}  & \multicolumn{2}{c}{Pyraformer} \\ \midrule
\multicolumn{1}{c|}{Dataset} & $T$                            & \multicolumn{1}{c}{MSE}           & \multicolumn{1}{c|}{MAE}            & MSE         & MAE         & MSE         & MAE            & MSE           & MAE            & MSE           & MAE             & MSE         & MAE            & MSE         & MAE                 & MSE         & MAE                \\ \midrule
\multicolumn{1}{c|}{\multirow{4}{*}{Traffic}}     & 96  & \textbf{0.356}$^*$ & \textbf{0.244}$^*$ & {\ul 0.367}         & {\ul 0.251}     & 0.410            & 0.282               & 0.473                & 0.293                 & 0.595                & 0.318                             & 0.576                & 0.359         & 0.597  & 0.371                            & 2.085                & 0.468                \\
\multicolumn{1}{c|}{}                             & 192 & \textbf{0.375}$^*$ & \textbf{0.251}$^*$ & {\ul 0.385}         & {\ul 0.259}     & 0.423            & 0.287               & 0.483                & 0.298                 & 0.615                & 0.326                             & 0.610                & 0.380         & 0.607  & 0.382                           & 0.867                & 0.467                \\
\multicolumn{1}{c|}{}                             & 336 & \textbf{0.386}$^*$ & \textbf{0.256}$^*$ & {\ul 0.398}         & {\ul 0.265}     & 0.436            & 0.296               & 0.491                & 0.303                 & 0.616                & 0.326                             & 0.608                & 0.375         & 0.623  & 0.387                          & 0.869                & 0.469                \\
\multicolumn{1}{c|}{}                             & 720 & \textbf{0.425}$^*$ & \textbf{0.279}$^*$ & {\ul 0.434}         & {\ul 0.287}     & 0.466            & 0.315               & 0.559                & 0.327                 & 0.655                & 0.353                             & 0.621                & 0.375         & 0.639  & 0.395                          & 0.881                & 0.473                \\ \midrule
\multicolumn{1}{c|}{\multirow{4}{*}{Electricity}} & 96 & \textbf{0.127}$^*$  & \textbf{0.222}$^*$  & {\ul 0.130}         & {\ul 0.222}     & 0.140            & 0.237               & 0.157                & 0.266                 & 0.178                & 0.284                             & 0.186               & 0.302         & 0.196  & 0.313                   & 0.386                & 0.449                \\
\multicolumn{1}{c|}{}                             & 192 & \textbf{0.144}$^*$  & \textbf{0.238}$^*$  & {\ul 0.148}         & {\ul 0.240}     & 0.153            & 0.249               & 0.175                & 0.287                 & 0.187                & 0.289                             & 0.197               & 0.311        & 0.211  & 0.324                    & 0.386                & 0.443                \\
\multicolumn{1}{c|}{}                             & 336 & \textbf{0.162}$^*$ & \textbf{0.256}$^*$ & {\ul 0.167}         & {\ul 0.261}     & 0.169            & 0.267               & 0.200                & 0.308                 & 0.208                & 0.307                             & 0.213                & 0.328         & 0.214  & 0.327                   & 0.378                & 0.443                \\
\multicolumn{1}{c|}{}                             & 720 & \textbf{0.199}$^*$ & \textbf{0.289}$^*$ & {\ul 0.202}         & {\ul 0.291}     & 0.203            & 0.301               & 0.228                & 0.338                 & 0.245                & 0.321                             & 0.233                & 0.344         & 0.236  & 0.342                   & 0.376                & 0.445                \\ \midrule
\multicolumn{1}{c|}{\multirow{4}{*}{Weather}}     & 96 & \textbf{0.150}$^*$   & \textbf{0.199}$^*$  & {\ul 0.152}         & {\ul 0.199}     & 0.176            & 0.237               & 0.178                & 0.249                 & 0.163                & 0.219                             & 0.238                & 0.314       & 0.249  & 0.329                          & 0.896                & 0.556                \\
\multicolumn{1}{c|}{}                             & 192 & \textbf{0.194}$^*$  & \textbf{0.240}$^*$  & {\ul 0.197}         & {\ul 0.243}     & 0.220            & 0.282               & 0.243                & 0.269                 & 0.211                & 0.259                             & 0.275                & 0.329       & 0.325  & 0.370                          & 0.622                & 0.624                \\
\multicolumn{1}{c|}{}                             & 336 & \textbf{0.246}$^*$  & \textbf{0.281}$^*$  & {\ul 0.249}         & {\ul 0.283}     & 0.265            & 0.319               & 0.278                & 0.338                 & 0.286                & 0.311                             & 0.339                & 0.377       & 0.351  & 0.391                          & 0.739                & 0.753                \\
\multicolumn{1}{c|}{}                             & 720 & \textbf{0.319}$^*$  & \textbf{0.333}$^*$  & {\ul 0.320}         & {\ul 0.335}     & 0.323            & 0.362               & 0.320                & 0.360                 & 0.359                & 0.363                             & 0.389                & 0.409       & 0.415  & 0.426                          & 1.004                & 0.934                \\ \midrule
\multicolumn{1}{c|}{\multirow{4}{*}{ETTh1}}       & 96  & \textbf{0.358}$^*$ & \textbf{0.390}$^*$  & {\ul 0.375}         & {\ul 0.399}     & {\ul 0.375}      & {\ul 0.399}         & 0.413                & 0.442                 & 0.421                & 0.440                             & 0.376                & 0.415        & 0.435  & 0.446                          & 0.664                & 0.612                \\
\multicolumn{1}{c|}{}                             & 192 & \textbf{0.396}$^*$ & \textbf{0.414}$^*$ & 0.414               & 0.421           & {\ul 0.405}      & {\ul 0.416}         & 0.451                & 0.462                 & 0.511                & 0.498                             & 0.423                & 0.446         & 0.456  & 0.457                         & 0.790                & 0.681                \\
\multicolumn{1}{c|}{}                             & 336 & \textbf{0.391}$^*$ & \textbf{0.420}$^*$ & {\ul 0.431}         & {\ul 0.436}     & 0.439            & 0.443               & 0.556                & 0.528                 & 0.484                & 0.478                             & 0.444                & 0.462         & 0.486  & 0.487                         & 0.891                & 0.738                \\
\multicolumn{1}{c|}{}                             & 720 & \textbf{0.430}$^*$ & \textbf{0.457}$^*$ & {\ul 0.449}         & {\ul 0.466}     & 0.472            & 0.490               & 0.658                & 0.607                 & 0.554                & 0.527                             & 0.469                & 0.492         & 0.515  & 0.517                         & 0.963                & 0.782                \\ \midrule
\multicolumn{1}{c|}{\multirow{4}{*}{ETTh2}}       & 96  & \textbf{0.257}$^*$  & \textbf{0.326}$^*$  & {\ul 0.274}         & {\ul 0.336}     & 0.289            & 0.353               & 0.303                & 0.364                 & 0.366                & 0.417                             & 0.332                & 0.374       & 0.332  & 0.368                         & 0.645                & 0.597                \\
\multicolumn{1}{c|}{}                             & 192 & \textbf{0.309}$^*$  & \textbf{0.361}$^*$  & {\ul 0.339}         & {\ul 0.379}     & 0.383            & 0.418               & 0.403                & 0.446                 & 0.426                & 0.447                             & 0.407                & 0.446       & 0.426  & 0.434                      & 0.788                & 0.683                \\
\multicolumn{1}{c|}{}                             & 336 & \textbf{0.302}$^*$ & \textbf{0.366}$^*$ & {\ul 0.331}         & {\ul 0.380}     & 0.448            & 0.465               & 0.603                & 0.550                 & 0.406                & 0.435                             & 0.400                & 0.447         & 0.477  & 0.479                    & 0.907                & 0.747                \\
\multicolumn{1}{c|}{}                             & 720 & \textbf{0.372}$^*$  & \textbf{0.416}$^*$  & {\ul 0.379}         & {\ul 0.422}     & 0.605            & 0.551               & 1.106                & 0.852                 & 0.427                & 0.457                             & 0.412                & 0.469       & 0.453  & 0.490                       & 0.963                & 0.783                \\ \midrule
\multicolumn{1}{c|}{\multirow{4}{*}{ETTm1}}       & 96  & \textbf{0.286}$^*$  & \textbf{0.338}$^*$  & {\ul 0.290}         & {\ul 0.342}     & 0.299            & 0.343               & 0.308                & 0.360                 & 0.356                & 0.385                             & 0.326                & 0.390       & 0.510  & 0.492                             & 0.543                & 0.510                \\
\multicolumn{1}{c|}{}                             & 192 & \textbf{0.327}$^*$  & {\ul 0.366}     & {\ul 0.332}         & 0.369           & 0.335            & \textbf{0.365}      & 0.343                & 0.384                 & 0.452                & 0.428                     & 0.365                & 0.415                   & 0.514  & 0.495                            & 0.557                & 0.537                \\
\multicolumn{1}{c|}{}                             & 336 & \textbf{0.362}$^*$  & {\ul 0.389}  & {\ul 0.366}         & {\ul 0.392}     & 0.369            & \textbf{0.386}               & 0.395                & 0.411                 & 0.419                & 0.425                     & 0.392                & 0.425             & 0.510  & 0.492                           & 0.754                & 0.655                \\
\multicolumn{1}{c|}{}                             & 720 & \textbf{0.414}$^*$ & \textbf{0.421} & {\ul 0.420}         & {\ul 0.424}     & 0.425            & \textbf{0.421}      & 0.427                & 0.434                 & 0.452                & 0.451                     & 0.446                & 0.458                     & 0.527  & 0.493                          & 0.908                & 0.724                \\ \midrule
\multicolumn{1}{c|}{\multirow{4}{*}{ETTm2}}       & 96  & \textbf{0.162}$^*$  & \textbf{0.251}$^*$ & {\ul 0.165}         & {\ul 0.255}     & 0.167            & 0.260               & 0.169                & 0.268                 & 0.188                & 0.276                     & 0.180                & 0.271                & 0.205  & 0.293                           & 0.435                & 0.507                \\
\multicolumn{1}{c|}{}                             & 192 & \textbf{0.220}   & \textbf{0.291}  & \textbf{0.220}      & {\ul 0.292}     & {\ul 0.224}      & 0.303               & 0.247                & 0.333                 & 0.242                & 0.310                     & 0.252                & 0.318                      & 0.278  & 0.336                            & 0.730                & 0.673                \\
\multicolumn{1}{c|}{}                             & 336 & \textbf{0.272}$^*$  & \textbf{0.326}$^*$  & {\ul 0.278}         & {\ul 0.329}     & 0.281            & 0.342               & 0.290                & 0.351                 & 0.300                & 0.346                     & 0.324                & 0.364               & 0.343  & 0.379                             & 1.201                & 0.845                \\
\multicolumn{1}{c|}{}                             & 720 & \textbf{0.358}$^*$ & \textbf{0.379}$^*$ & {\ul 0.367}         & {\ul 0.385}     & 0.397            & 0.421               & 0.417                & 0.434                 & 0.391                & 0.403                             & 0.410                & 0.420         & 0.414  & 0.419                            & 3.625                & 1.451                \\ \midrule
% \multicolumn{1}{c|}{\multirow{4}{*}{ILI}}       & 24   & \textbf{1.499*} & \textbf{0.790*} & {\ul 1.522}   & {\ul 0.814}          & 2.215             & 1.081              &      2.345       &    1.043         &     2.157         &     0.978   & 2.624 & 1.095& 2.906& 1.182 & 1.420& 2.012        \\
% \multicolumn{1}{c|}{}                             & 36 & \textbf{1.413*} & \textbf{0.789*} & {\ul 1.430}   & {\ul 0.834}         &1.963 & 0.963  &        2.330      &     1.001        &        2.318       &    1.031      &  2.516 & 1.021 & 2.585 & 1.038 & 7.394 &2.031        \\
% \multicolumn{1}{c|}{}                             & 48 & \textbf{1.605*} & {\ul 0.877} & {\ul 1.673}   & \textbf{0.854*}   & 2.130         & 1.024        &   2.386          &     1.051        &     2.121          &   1.005   & 2.505 & 1.041 & 3.024 & 1.145 & 7.551 & 2.057        \\
% \multicolumn{1}{c|}{}                             & 60 & \textbf{1.499*} & \textbf{0.814*} & {\ul 1.529}   & {\ul 0.862}    & 2.368         & 1.096        &    2.616         &    1.131         &       1.975        &      0.975      &  2.742 & 1.122 & 2.761 & 1.114 & 7.662 & 2.100     \\ \midrule

\multicolumn{2}{c|}{\#Rank-1st (total=28)}        & \multicolumn{1}{c}{28} & \multicolumn{1}{c|}{26} &  1         &  0   & 0            & 3               &  0               & 0 & 0                      & 0 & 0 & 0 &  0 & 0        &  0 & 0         \\ \bottomrule
\end{tabular}
\label{table:mtst336}
\end{table*}
\subsubsection{Baselines and Experimental Setup}

We evaluate our proposed model in comparison with the \emph{state-of-the-art} (SOTA) time-series transformer -- {PatchTST}~\cite{nie2023TimeSeriesWorth};
three multiscale time-series models -- {MICN}~\cite{wang2023MicnMultiscaleLocal}, {TimesNet}~\cite{wu2022TimesnetTemporal2dVariation}, and {Pyraformer}~\cite{liu2021PyraformerLow-complexityPyramidal};
as well as a simple and competitive linear baseline -- DLinear~\cite{zeng2023AreTransformersEffective}.
We also include two earlier timestamp-based time-series transformers for reference -- Fedformer~\cite{zhou2022FEDformerFrequencyEnhanced} and Autoformer~\cite{wu2021AutoformerDecompositionTransformers}.

For fair comparison, 
we follow the experimental setup in \citet{nie2023TimeSeriesWorth} so that the prediction horizon $T{\in}\{96, 192, 336, 720\}$ and the look-back window $L{=}336$.
We report the baseline results from \citet{nie2023TimeSeriesWorth}\footnotemark \footnotetext{The results for Fedformer, Autoformer and Pyraformer are the best on each dataset from multiple look-back windows $L\in \{24, 48, 96, 192, 336, 720\}$.} except for TimesNet and MICN,
for which we reproduce 
their results with $L{=}336$ based on the officially released code.
Since Scaleformer~\cite{shabani2023Scaleformer} has multiple variants, and the best-performing variant differs according to the dataset under study, we do not include it in the comparison. We provide a comparison with Scaleformer in the Appendix (Table~\ref{tab:L96}). Note that in all experiments, models are trained and evaluated for each prediction horizon independently.

\subsubsection{Hyperparameters}

In all our experiments, we use instance-normalization and denormalization~\cite{kim2022ReversibleInstanceNormalization} on the input and prediction, respectively. 
In each MTST layer, the transformer layer contains scaled dot-product attention and batch normalization~\cite{ioffe2015BatchNormalizationAccelerating}, with the use of relative PE.
The model is trained using the Adam optimizer~\cite{kingma2015Adam} to minimize the MSE loss over the training set. The detailed hyperparameter configuration of MTST associated with the model architecture and the training process for each dataset/horizon is provided in Section~\ref{subsec:hyperparams} of the Appendix.

\subsubsection{Experimental Results}
\label{sec:main-res}
Table~\ref{tab:L336} reports the experimental results for seven benchmarks.
The performance is measured by MSE and MAE; the best and second-best results for each case (dataset, horizon, and metric) are highlighted in bold and underlined, respectively. We conduct the Wilcoxon signed rank test~\cite{wilcoxon1945IndividualComparisonsRanking} with bootstrap sampling with significance level 5\% on the top-2 results; we use $^*$ to denote a significant difference.

Our proposed model, MTST, achieves SOTA performances in all cases (7 datasets, 4 horizons, and 2 metrics). 
MTST outperforms, with statistical significance, the previous SOTA patch-based TST -- PatchTST -- on 27 out of 28 cases for the MSE metric.

We rank the algorithms in Table~\ref{tab:L336} based on their MSE and order them based on their average rank across seven datasets and four prediction horizons. 
Figure~\ref{fig:rank} shows the boxplot of rank. We observe that the proposed MTST achieves the best average rank and lowest variability across all settings.

As a multi-scale convolution-based approach, MICN demonstrates competitive performance for small prediction horizons $T$,
but encounters more severe performance degradation for larger $T$. 
This matches the observations in previous works that convolution-based models struggle to capture long-range dependencies. This motivates the development of transformer-based models for long-term forecasting.
Notably, DLinear, a simple linear model, outperforms earlier timestamp-based TSTs (Fedformer, Autoformer, Pyraformer) with a noticeable performance gap, 
indicating the ineffectiveness of the timestamp-based TSTs and the need to develop patch-based TSTs.

\subsection{Ablation Study and Analysis Experiment}

\subsubsection{Ablation: Multi-Resolution }

We conduct an ablation experiment to study the usefulness of the multi-resolution representation on 5 datasets: ETTh1, ETTh2, ETTm1, Weather, and Traffic, under the same setup of the main experiment.
To verify the importance of each resolution, 
we compare MTST against two variants: \emph{w/o Low-RES.} stands for the variant without the lowest-resolution representations (i.e., the branch with the largest-sized patch), which mainly contributes to low-frequency temporal patterns; 
\emph{w/o High-RES.} denotes the variant removing the highest-resolution representations (i.e., the branch with the smallest-sized patches), which mainly aims to capture high-frequency temporal patterns.

From Table~\ref{tab:resolution}, we can see that removing either resolution generally leads to performance degradation.
However, removing the highest-resolution branch results in slightly better performance in a few cases for small datasets (e.g., ETTm1).
This is potentially due to overfitting, which usually focuses on high-frequency components.

\begin{table}[t!]
\caption{Ablation study on multi-branch architecture.}
\label{tab:resolution}
\scriptsize
\setlength{\tabcolsep}{1pt}
\begin{tabular}{cc|cc|cc|cc}
\toprule
\multicolumn{2}{c|}{Models}                         & \multicolumn{2}{c|}{MTST (Base)} & \multicolumn{2}{c|}{w/o Low-RES.} & \multicolumn{2}{c}{w/o High-RES.} \\ \midrule
\multicolumn{1}{c|}{Dataset} &  $T$                          & MSE         & MAE         & MSE               & MAE               & MSE                & MAE               \\ \midrule
\multicolumn{1}{c|}{\multirow{4}{*}{ETTh1}}   & 96  &\textbf{0.358}       & \textbf{0.390}      & 0.373             & 0.402             & 0.372              & 0.400             \\
\multicolumn{1}{c|}{}                         & 192 & \textbf{0.396}       & \textbf{0.414}       & 0.397             & 0.418             & 0.399              & 0.424             \\
\multicolumn{1}{c|}{}                         & 336 & \textbf{0.391}       & \textbf{0.420}       & 0.397             & 0.428             & 0.399              & 0.424             \\
\multicolumn{1}{c|}{}                         & 720 & \textbf{0.430}       & \textbf{0.457}       & 0.435             & 0.460             & \textbf{0.430}              & \textbf{0.457}             \\ \midrule
\multicolumn{1}{c|}{\multirow{4}{*}{ETTh2}}   & 96  & \textbf{0.257}       & \textbf{0.326}       & 0.260             & 0.329             & 0.266              & 0.335             \\
\multicolumn{1}{c|}{}                         & 192 & \textbf{0.309}       & \textbf{0.361}       & 0.311             & 0.364             & 0.317              & 0.370             \\
\multicolumn{1}{c|}{}                         & 336 & \textbf{0.302}       & \textbf{0.366}       & 0.304             & 0.369             & 0.311              & 0.376             \\
\multicolumn{1}{c|}{}                         & 720 & \textbf{0.372}       & \textbf{0.416}       & 0.373             & 0.417             & 0.380              & 0.422             \\ \midrule
\multicolumn{1}{c|}{\multirow{4}{*}{ETTm1}}   & 96  & 0.286       & \textbf{0.338}       & 0.290             & 0.341             & \textbf{0.285}              & \textbf{0.338}             \\
\multicolumn{1}{c|}{}                         & 192 & 0.327       & 0.366       & 0.334             & 0.372             & \textbf{0.322}              & \textbf{0.364}             \\
\multicolumn{1}{c|}{}                         & 336 & \textbf{0.362}       & 0.389       & 0.367             & 0.389             & 0.365              & \textbf{0.386}             \\
\multicolumn{1}{c|}{}                         & 720 & \textbf{0.414}       & \textbf{0.421}       & 0.422             & \textbf{0.421}             & 0.419              & 0.423             \\ \midrule
\multicolumn{1}{c|}{\multirow{4}{*}{Weather}} & 96  & \textbf{0.150}       & \textbf{0.199}       & 0.151             & \textbf{0.199}             & 0.151              & \textbf{0.199}             \\
\multicolumn{1}{c|}{}                         & 192 & \textbf{0.194}       & \textbf{0.240}       & 0.196             & \textbf{0.240}              & 0.196              & 0.243             \\
\multicolumn{1}{c|}{}                         & 336 & \textbf{0.246}       & 0.281       & \textbf{0.246}             & \textbf{0.280}            & 0.247              & \textbf{0.280}              \\
\multicolumn{1}{c|}{}                         & 720 & \textbf{0.319}       & \textbf{0.333}       & 0.322             & 0.335             & 0.324              & 0.335             \\ \midrule
\multicolumn{1}{c|}{\multirow{4}{*}{Traffic}} & 96  & \textbf{0.356} & \textbf{0.244}          & 0.357             & \textbf{0.244}    & 0.362              & 0.249             \\
\multicolumn{1}{c|}{}                         & 192 & \textbf{0.375} & \textbf{0.251} & 0.379             & 0.253             & 0.378              & 0.253             \\
\multicolumn{1}{c|}{}                         & 336 & \textbf{0.386} & 0.258          & 0.388             & \textbf{0.257}    & 0.391              & 0.261             \\
\multicolumn{1}{c|}{}                         & 720 & \textbf{0.425} & 0.281          & \textbf{0.425}    & \textbf{0.279}    & 0.427              & 0.283             \\ \midrule
\multicolumn{2}{c|}{\#Rank-1 (total=20)}    & 18       & 15       &  2             & 7             & 3              & 6         \\ \bottomrule
\end{tabular}
\end{table}

\begin{table}[t!] 
\scriptsize
\centering
\caption{Ablation study on positional encoding.}
\label{table:PE}
\setlength{\tabcolsep}{2.4pt}
\begin{tabular}{cc|cc|cc|cc}
\toprule
\multicolumn{2}{c|}{MTST w/}                               & \multicolumn{2}{c|}{RPE (ours)} & \multicolumn{2}{c|}{SinAPE}        & \multicolumn{2}{c}{LenaredAPE}       \\ \midrule
\multicolumn{1}{c|}{Dataset} & $T$                            & MSE         & MAE            & MSE                & MAE          & MSE               & MAE                           \\ \midrule
\multicolumn{1}{c|}{\multirow{4}{*}{ETTh1}}   & 96       & \textbf{0.358}      & \textbf{0.390}          & 0.362              & 0.392        & 0.363             & 0.394                         \\
\multicolumn{1}{c|}{}                         & 192      & \textbf{0.396}       & \textbf{0.414}          & 0.400              & 0.417        & 0.400             & 0.420                         \\
\multicolumn{1}{c|}{}                         & 336      & 0.391       & 0.420          & \textbf{0.389}              & \textbf{0.419}        & 0.390             & \textbf{0.419}                        \\
\multicolumn{1}{c|}{}                         & 720      & \textbf{0.430}       & \textbf{0.457}          & 0.440              & 0.464        & 0.435             & 0.462                         \\ \midrule
\multicolumn{1}{c|}{\multirow{4}{*}{ETTh2}}   & 96       & \textbf{0.257}       & \textbf{0.326}          & 0.258              & 0.328        & 0.258             & 0.328                         \\
\multicolumn{1}{c|}{}                         & 192      & \textbf{0.309}       & \textbf{0.361}          & \textbf{0.309}              & 0.363        & 0.311             & 0.365                         \\
\multicolumn{1}{c|}{}                         & 336      & \textbf{0.302}       & \textbf{0.366}          & 0.305              & 0.371        & 0.306             & 0.373                         \\
\multicolumn{1}{c|}{}                         & 720      & \textbf{0.372}       & \textbf{0.416}          & 0.377              & 0.422        & 0.375             & 0.420                         \\ \midrule
\multicolumn{1}{c|}{\multirow{4}{*}{ETTm1}}   & 96       & \textbf{0.286}       & \textbf{0.338}          & 0.297              & 0.346        & 0.290             & 0.342                         \\
\multicolumn{1}{c|}{}                         & 192      & \textbf{0.327}       & \textbf{0.366}          & 0.332              & 0.370        & 0.330             & 0.370                         \\
\multicolumn{1}{c|}{}                         & 336      & \textbf{0.362}       & 0.389          & 0.369              & 0.390        & 0.366             & \textbf{0.388}                         \\
\multicolumn{1}{c|}{}                         & 720      & \textbf{0.414}       & \textbf{0.421}          & 0.419              & 0.423        & 0.420             & 0.425                         \\ \midrule
\multicolumn{1}{c|}{\multirow{4}{*}{Weather}} & 96       & \textbf{0.150}       & 0.199          & \textbf{0.150}              & \textbf{0.197}        & \textbf{0.150}             & 0.198                         \\
\multicolumn{1}{c|}{}                         & 192      & \textbf{0.194}       & \textbf{0.240}          & \textbf{0.194}              & \textbf{0.240}        & \textbf{0.194}             & \textbf{0.240}                         \\
\multicolumn{1}{c|}{}                         & 336      & \textbf{0.246}       & \textbf{0.281}          & 0.247              & \textbf{0.281}        & 0.247             & 0.282                         \\
\multicolumn{1}{c|}{}                         & 720      & \textbf{0.319}       & \textbf{0.333}          & 0.326              & 0.339        & 0.320             & \textbf{0.333}                         \\ \midrule

\multicolumn{1}{c|}{\multirow{4}{*}{Traffic}} & 96  & \textbf{0.356} & \textbf{0.244} & 0.361             & 0.247             & 0.360              & 0.246             \\
\multicolumn{1}{c|}{}                         & 192 & \textbf{0.375} & \textbf{0.251} & 0.381             & 0.257             & 0.381              & 0.256             \\
\multicolumn{1}{c|}{}                         & 336 & \textbf{0.386} & \textbf{0.258} & 0.393             & 0.263             & 0.397              & 0.266             \\
\multicolumn{1}{c|}{}                         & 720 & \textbf{0.425} & \textbf{0.281} & 0.432             & 0.288             & 0.431              & 0.286             \\  \midrule

\multicolumn{2}{c|}{\#Rank-1 (total=20)}           &    19   &   17 & 4              & 4          & 2               & 4                            \\
\bottomrule
\end{tabular}
\end{table}

\begin{figure}[h!]
\centering
% \hspace{-1em}
\vspace{1.5em}
\includegraphics[width=0.49\textwidth]{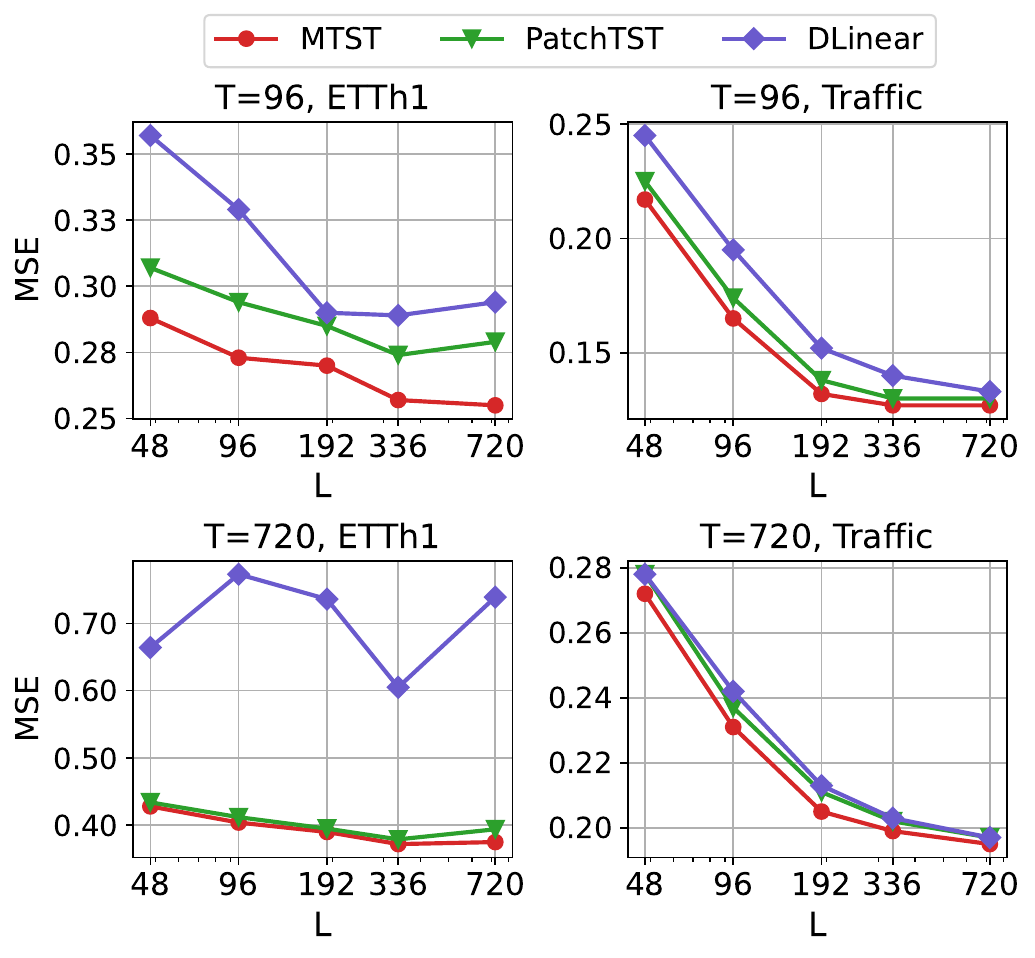}
\caption{MSE at different look-back windows on ETTh1 and Traffic datasets, $L{\in}\{48, 96, 192, 336, 720\}$ and $T{=}96 \text{ and } 720$. 
}
\label{fig:lookback}
\end{figure}

\begin{figure*}[htbp]
\centering
    \begin{minipage}{\textwidth}
    \centering
        \includegraphics[width=0.985\textwidth]{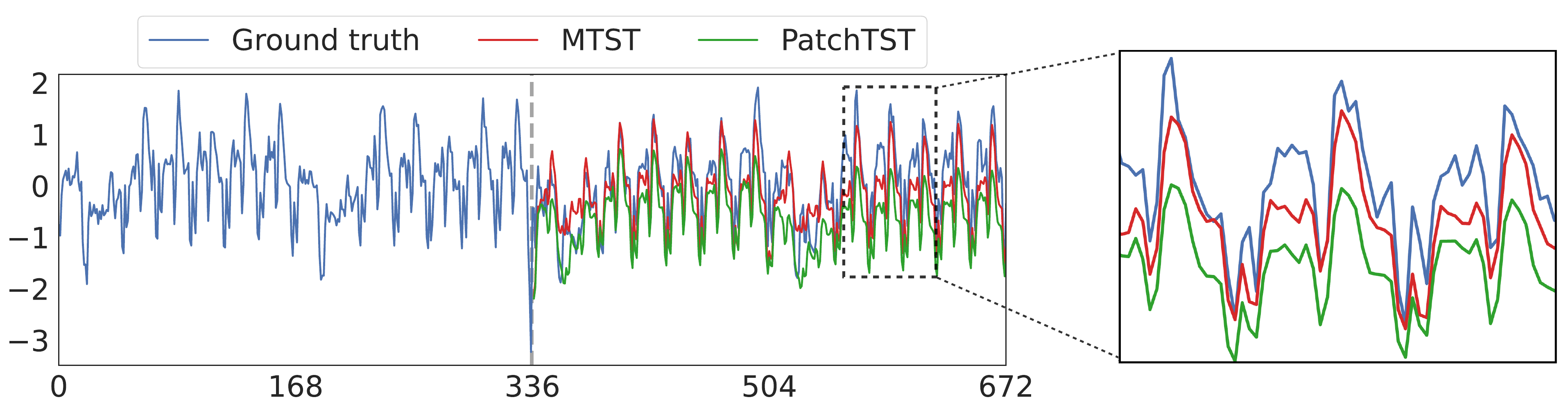}
        % \subcaption{$310$-th variate on $4050$-th example on \text{Electricity}.}
        \subcaption{$318$-th variate on $4646$-th example on \text{Electricity}.}
        \label{fig:visual_electricity}
    \end{minipage} 
    \hfill
    \vspace{1em}
    \begin{minipage}{\textwidth}
        \centering
        \includegraphics[width=0.985\textwidth]{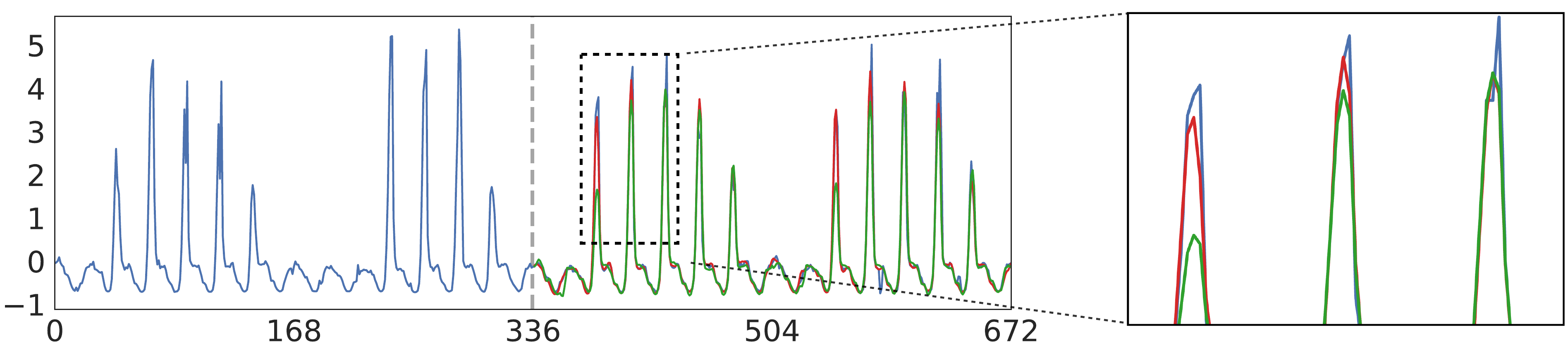}
        \subcaption{$4$-th variate on $807$-th example on \text{Traffic}.}
        \label{fig:visual_traffic}
    \end{minipage}
    \caption{Visualization of forecasts for PatchTST and MTST ($L{=}336$ and $T{=}336$).}
    \label{fig:visual_example}
\end{figure*}

\subsubsection{Ablation: Positional Encoding}

Positional encoding is known to be  an important component in transformer design. However, previous TSTs employ absolute positional encoding by default.
Therefore, we conduct an ablation experiment to study whether the choice of positional encoding for time-series forecasting is important.
Using our design of MTST with relative positional encoding (RPE) as the base model, we compare with the widely used fixed absolute positional encoding (SinAPE), introduced by \citet{vaswani2017AttentionAllYou}, as well as the learned absolute positional encoding (LearnedAPE) used in PatchTST~\cite{nie2023TimeSeriesWorth}.
The APE is injected into the token representations after patch-based tokenization in each branch of every layer.

From the results of the PE ablation study (shown in Table~\ref{table:PE}),
the incorporation of the relative PE is beneficial to the outcome of forecasting for most cases.
The observation also matches the finding from previous works in other domains that RPE performs better than APE when periodicities are important~\cite{huang2019MusicTransformerGenerating}.

% \begin{figure*}[htbp]
% \centering
% \hspace{-1em}
% \includegraphics[width=\textwidth]{Figure/others/lookback_v5.pdf}
% \caption{MSE v.s. look-back window $L$ on ETTh1, ETTh2, Electricity, and Traffic datasets. 
% }
% \label{fig:lookback}
% \end{figure*}

% \begin{figure*}[t!]
% \centering
%     \begin{minipage}{\textwidth}
%     \centering
%         % \includegraphics[width=\textwidth]{Figure/examples/Electricity_pl336_var318_sample4646.pdf}
%         \includegraphics[width=\textwidth]{Figure/examples/new_zoomin/zoom-in-electricity.pdf}
%         % \subcaption{$310$-th variate on $4050$-th example on \text{Electricity}.}
%         \subcaption{$318$-th variate on $4646$-th example on \text{Electricity}.}
%         \label{fig:visual_electricity}
%     \end{minipage} \hfill
%     \begin{minipage}{\textwidth}
%         \centering
%         % \includegraphics[width=\textwidth]{Figure/examples/traffic_pl336_var4_sample807.pdf}
%         \includegraphics[width=\textwidth]{Figure/examples/new_zoomin/zoom-in-traffic.pdf}
%         \subcaption{$4$-th variate on $807$-th example on \text{Traffic}.}
%         \label{fig:visual_traffic}
%     \end{minipage}
%     \caption{Visualization of forecasts for PatchTST and MTST ($L=336$ and $T=336$).}
%     \label{fig:visual_example}
% \end{figure*}

\subsubsection{Analysis: Look-back Window}

In the main experiment (Table~\ref{tab:L336}), MTST reaches state-of-the-art performance with $L=336$.
We now conduct a more in-depth study into MTST's behavior for different look-back windows.
For the \text{ETTh1} and \text{Traffic} datasets, we train and evaluate our model with $L\in \{48, 96, 192, 336, 720\}$ independently, and compare with the second and third best models (PatchTST and DLinear).

We visualize the results with $T=96$ and $720$ in Figure~\ref{fig:lookback}. 
Although all algorithms show improved performance with increasing the look-back window in most cases, the advantages of MTST over PatchTST and DLinear are retained over nearly all look-back windows for $T=96 \text{ and } 720$. 
% showcasing that the multi-resolution design can improve patch-based TSTs across a wide range of look-back window durations.
% Additionally, 
% suggesting that a strength of MTST is its ability to capture the long-term temporal dependencies while retaining its capability to represent higher-frequency temporal structure. 

% \begin{figure}[h!]
% \centering
% % \hspace{-1em}
% \vspace{1.5em}
% \includegraphics[width=0.49\textwidth]{Figure/others/lookback_final.eps}
% \caption{MSE at different look-back windows on ETTh1 and Traffic datasets, $L{\in}\{48, 96, 192, 336, 720\}$ and $T{=}96 \text{ and } 720$. 
% }
% \label{fig:lookback}
% \end{figure}

\subsubsection{Analysis: Qualitative Comparisons}
To complement the quantitative results, Figure~\ref{fig:visual_example} provides visualizations of two randomly chosen test examples from the \text{Electricity} and \text{Traffic} datasets. 
% This facilitates a better understanding of the advantage contributed by the multi-resolution representation.
We plot the ground truth in the look-back window $L=336$ and prediction horizon $T=336$, as well as the predictions from MTST. 
We include PatchTST as a representative single-scale patch-based TST. 
More visualizations are included in the Appendix (Section~\ref{sec:app:visual}).
From Figure~\ref{fig:visual_example}, we observe that although both PatchTST and MTST
can capture the overall temporal patterns, MTST's forecasts are superior to that of PatchTST.
Specifically, in Figure~\ref{fig:visual_electricity}, MTST is less impacted by the abnormally deep trough near the end of the look-back window, potentially owing to the low-resolution branch, and consequently performs better on predicting the overall trend.
In Figure~\ref{fig:visual_traffic}, 
PatchTST is more impacted by the low amplitude of the first high-peak in the second weekly cycle in the look-back window,
while MTST can forecast the amplitude of the first high peak better, potentially due to the finer prediction of the high-resolution branch.

% Specifically, in Figure~\ref{fig:visual_electricity}, MTST is less impacted by the abnormally deep trough near the end of the look-back window and performs better on predicting the overall trend.
% In Figure~\ref{fig:visual_traffic}, MTST can predict the amplitude of the first high peak better compared to PatchTST.

% From Figure~\ref{fig:visual_electricity},  although both PatchTST and MTST can capture the seasonalities, MTST performs better in predicting the trend of the future data. 
% % \MJC{What do you mean by ``bias''?}\mlh{The global trend.}
% One potential reason is that the lower-resolution branch can focus on the trend and long-term seasonality, while the higher-resolution branch targets the more fine-grained structure. \MJC{This seems flimsy too - do the branch-level predictions support this claim at all?} In Figure~\ref{fig:visual_traffic}, both PatchTST and MTST can capture the overall pattern of the time-series.
% However, MTST can recover the scale of the spikes better than PatchTST, which is potentially due to the benefits of the higher-resolution branch in MTST.
% \MJC{Why - this claim seems very flimsy}
% \MJC{Can you perhaps investigate this further - MTST seems to be doing a nice job in understanding that the previous Monday may have been a holiday - of course, it doesn't really understand this, but it has two examples, one where Monday is large and another where it is small. Can you try to figure out what the Monday prediction is averaging? }  

\section{CONCLUSION}
\label{sec:conclusion}
In this paper, we incorporate multi-scale analysis into patch-based time-series transformers,
and accordingly
propose a novel framework, termed Multi-resolution Time-Series Transformer (MTST), for forecasting.
Contrary to up/down-sampling techniques used in previous work, 
we use multiple patch-based tokenizations with different patch-sizes together with the multi-branch transformer architecture in each MTST layer,
which enables flexibly modeling of temporal patterns of different scales.
% The proposed architecture has multiple branches with different patch sizes for flexible modeling of temporal pattern of different scales.
Additionally, we propose to incorporate relative positional encoding in our design, which is well-matched to the seasonality characteristics present in many real-world datasets.
Extensive experimental results on seven multivariate time-series datasets demonstrate that the proposed MTST outperforms the previous \emph{state-of-the-art} approaches.
The example visualization of the prediction from each branch on real-world datasets also provides validation for our hypothesis on multi-scale learning.
Furthermore, we have conducted ablation studies on multiple resolutions and positional encoding, which justify our design choices. 
% To address a challenge in our current model, where patch sizes are manually adjusted for each resolution, future models could be improved by incorporating a mechanism that automatically identifies the optimal patch sizes during the training process.
% Nonetheless, MTST is not the final chapter for improving patch-based time-series transformers:
% future work can address limitations of our work, such as finding the good choices of patch-sizes in MTSTs, as well as in previous works,
% requires prior knowledge of the data or heavy hyperparameter-search.

\section{ACKNOWLEDGEMENT}
% This work was partially supported by the Department of
% National Defence’s Innovation for Defence Excellence and
% Security (IDEaS) program, Canada. 
We acknowledge
the support of the Natural Sciences and Engineering Research Council of Canada (NSERC), [funding reference number 260250].\\
Cette recherche a été financée par le Conseil de recherches en sciences naturelles et en génie du Canada (CRSNG), [numéro de référence 260250].

\bibliography{ref_in_use}

\onecolumn

% \aistatstitle{Supplementary Material for Multi-resolution Time-Series Transformer for Long-term Forecasting}
\section*{\center SUPPLEMENTARY MATERIAL}

\vspace{0.5cm}
\renewcommand\thesection{\Alph{section}}
\renewcommand\thesubsection{\thesection.\arabic{subsection}}

\setcounter{section}{0}
\section{IMPLEMENTATION DETAILS}

\subsection{Detailed Architecture of Transformer Layers in MTST}
\label{subsec:layer_details}
As written in Section~\ref{subsec:transformer} of the main paper, here we describe the full architectural details of the transformer layer. The output of the self-attention operation $\widetilde{\bZ} {=} \func{Attn}(\widetilde{\bY})$ is further processed  with a particular combination of residual connections,  batch normalizations~\cite{ioffe2015BatchNormalizationAccelerating}, and a two-layer feed-forward network $\func{FFN}_{\theta}:\sR^{D}\rightarrow\sR^{D}$ as follows:
\begin{align}
\widetilde{\bZ}' &= \func{BN}\big(\widetilde{\bY} + \func{Attn}(\widetilde{\bY})\big)\,,\\
\widetilde{\bZ}'' &=\func{BN}\big(\widetilde{\bZ}' + \func{FFN}_{\theta}(\widetilde{\bZ}')\big)\,.
\end{align}
Here, $\func{BN}(\cdot)$ stands for batch-normalization, and $\theta$ denotes the learnable weights and biases inside the $\func{FFN}$. 
Subsequently, $\widetilde{\bZ}''$-s from different branches are fused together as described in Section~\ref{subsec:fuse}.

\subsection{Computational Complexity}
\label{subsec:complexity}
As a transformer-based model, the asymptotic computational complexity of the operations in the $n$-th layer of MTST is $O(J_{b'_n}^2)$, where $J_{b'_n} {=} \lceil (d_{n-1}{-}P_{b'_n})/S_{b'_n} \rceil + 1$ is the number of tokens in the branch with the finest resolution. 
In other words, MTST has the same asymptotic computational complexity as PatchTST when $J_{b'_n}$ equals the number of tokens in PatchTST.
As $J_{b'_n} << L$, the computational requirement of MTST scales favorably compared to most timestamp-based TSTs with $O(L^2)$ complexity.

% \textbf{R1.4 Runtime:} As demonstrated in Sec.~1.2 in Appendix, MTST reaches its optimal performance with a shallower architecture compared to PatchTST based on the hyperparameter search. Therefore, despite having multiple branches, MTST has a similar training time compared to PatchTST. 

From the hyperparameter search, we empirically observe that MTST reaches its optimal performance with shallower architectures compared to PatchTST, as shown in Table~\ref{table:MTST_hp}.
Therefore, despite having multiple branches, the training time of MTST is similar to that of PatchTST.
% e.g. 249 sec/epoch (MTST) v.s. 250 sec/epoch (PatchTST) on Electricity dataset ($L=336, T=96$) 
Examples from 2 datasets with a single NVIDIA V100 GPU and 72 threads of Intel(R) Xeon(R) Gold 6140 CPU @ 2.30GHz are shown in Table~\ref{tab:time}. 
All our experiments were conducted with these hardware devices.

\begin{table}[htbp]
\begin{minipage}[t]{\linewidth}
\centering
\scriptsize
\caption{Training time $(L{=}336, T{=}96)$.}
\label{tab:time}
\centering
\begin{tabular}{c|cc|cc}
\toprule
Dataset         & \multicolumn{2}{c|}{Traffic}                                                         & \multicolumn{2}{c}{ETTm2}                            \\ \midrule
Model           & \multicolumn{1}{c|}{MTST}                           & PatchTST                       & \multicolumn{1}{c|}{MTST}           & PatchTST       \\ \midrule
% \#Parameter     & \multicolumn{1}{c|}{}                               &                                & \multicolumn{1}{c|}{3,388,784}      & 921,184        \\ \hline
Time(s) / Epoch & \multicolumn{1}{c|}{ 513.916} & {412.918} & \multicolumn{1}{c|}{69.871}         & 62.876         \\ 
\# Epoch        & \multicolumn{1}{c|}{68}                             & 77                             & \multicolumn{1}{c|}{28}             & 28             \\ 
Total time(h)   & \multicolumn{1}{c|}{9.707}                 & 8.832                 & \multicolumn{1}{c|}{0.543} & 0.489 \\ \bottomrule
\end{tabular}
\end{minipage} 
\end{table}

\section{EXPERIMENTAL DETAILS}
\label{subsec:exp_setting}

\subsection{Baseline settings}
The default look-back windows for different baseline models, as identified in the original papers and the code, can differ from each other.
For Transformer-based models, the default look-back window is $L=96$. 
TimesNet and MICN follow the same setting for a fair comparison.  

However, this can possibly lead to a poorer performance than if the algorithms are provided with a longer horizon.
A short 96 time-stamp look-back window is often inadequate for forecasting over longer horizons of more than 300 timestamps.
For example, MICN shows that the overall prediction performance gradually improves as the size of the look-back window increases. Therefore, we re-run TimesNet and MICN with a look back window $L=336$, which is the  default experimental setting used by PatchTST and DLinear.
We also conduct the experiments for MTST under the same look-back window for fair comparison.

In MICN, the principal hyperparameters are $stride {=} kernel$. We use the same setting when increasing the $L$ from 96 to 336, selecting the stride and kernel from $\{12, 16\}$. TimesNet selects the top-$k$ amplitude values and obtains the most significant frequencies $\{f_1, ..., f_k\}$ with amplitudes $\{A_{f_1}, ..., A_{f_k}\}$. Based on the calculation, the input is derived using $k$ different periods. We follow the setting for $L{=}96$ with $k{=}5$, indicating 5 different scales.

\subsection{Hyperparameters}
\label{subsec:hyperparams}
% The Multi-head attention block includes BatchNorm layers and a feed-forward network with residual connections as shown in Figure 2.
For fair comparison, most hyperparameter values are borrowed from the experimental study reported by~\citet{nie2023TimeSeriesWorth}. At each branch in each layer, the transformer layer consists of an attention mechanism with 16 heads, and a 2-layer feed-forward network with a hidden dimension of 256 and an output dimension of 128. We only perform grid search for the number of layers, $N{\in}\{1,2,3\}$, the number of branches, $B{\in}\{2,3,4\}$, and the patch-sizes in different branches $P_{b}{\in}\{4, 8, 16, 24, 36, 48, 64, 96\}$. The stride length is set to a default value of $S = P/2$.
% \mlh{Usually, people say that a transformer block consists of a attention block and a FFN block}
% At each branch in each layer, the transformer layer consists of an attention mechanism with 16 heads, and the 2-layer feed-forward network with hidden dimension of 256 and output dimension of 128.
% GELU activation function, which has hidden dimension $H=256$ and output dimension $D=128$.
The detailed hyperparameter configurations of MTST for all datasets are shown in Table~\ref{table:MTST_hp}. 
% We set the corresponding stride length as the half of the Patch length $S = P/2 $ for each branch. 
The model parameters are learned by minimizing the MSE of the forecasts on the training set using the Adam optimizer~\cite{kingma2015Adam}.

% with an initial learning rate ranging at $1e^{-4}$. 

\begin{table}[htbp]
\centering
\caption{Hyperparameters of MTST}
\scriptsize
\begin{tabular}{cc|ccccccc}
\toprule
\multicolumn{2}{c|}{Dataset}                                            & Traffic       & ELC           & Weather     & ETTh1                       & ETTh2                        & ETTm1                        & ETTm2                       \\ \midrule
\multicolumn{2}{c|}{Layer number N}                                     & 1             & 1             & 2           & 2                           & 1                            & 2                            & 2                            \\ \midrule
\multicolumn{2}{c|}{Branch Number B}                                    & 3             & 3             & 2           & 2                           & 2                            & 2                            & 2                            \\ \midrule
\multicolumn{1}{c|}{\multirow{2}{*}{Patch length P}} & T = \{96, 192\}  & {[}8,16,48{]} & {[}8,16,48{]} & {[}24,96{]} & \multirow{2}{*}{{[}8,16{]}} & \multirow{2}{*}{{[}16,96{]}} & \multirow{2}{*}{{[}16,96{]}} & \multirow{2}{*}{{[}16,96{]}} \\
\multicolumn{1}{c|}{}                                & T = \{336, 720\} & {[}8,32,96{]} & {[}8,32,96{]} & {[}16,96{]} &                             &                              &                              &                              \\ \midrule
\multicolumn{1}{c|}{\multirow{2}{*}{Stride   length S}} & T = \{96, 192\}  & {[}4,8,24{]}  & {[}4,8,24{]}  & {[}12,48{]} & \multirow{2}{*}{{[}4,8{]}}  & \multirow{2}{*}{{[}8,48{]}}  & \multirow{2}{*}{{[}8,48{]}}  & \multirow{2}{*}{{[}8,48{]}}  \\
\multicolumn{1}{c|}{}                                   & T = \{336, 720\} & {[}4,16,48{]} & {[}4,16,48{]} & {[}8,48{]}  &                             &                              &                              &                              \\ \midrule
\multicolumn{2}{c|}{Feed forward Dropout}                               & 0.2           & 0.2           & 0.2         & 0.3                         & 0.3                          & 0.2                          & 0.2                          \\ \midrule
\multicolumn{2}{c|}{Fusing layer Dropout}                               & 0             & 0             & 0           & 0.1                         & 0.3                          & 0                            & 0                            \\ \midrule
\multicolumn{2}{c|}{Batch size}                                         & 10            & 32            & 128         & 256                         & 256                          & 128                          & 128                          \\ \midrule
\multicolumn{2}{c|}{Initial leraning   rate}                            & $5e^{-4}$        & $5e^{-4}$         & $5e^{-4}$       & $1e^{-4}$                      & $1e^{-4}$                        & $1e^{-4}$                        & $1e^{-4}$                       \\ \bottomrule
\end{tabular}
\label{table:MTST_hp}
\end{table}

\begin{table}[!ht]
    \centering
    \scriptsize
    \caption{Multivariate forecasting results with different random seeds in MTST and PatchTST at $L{=}336$.}
    \label{tab:result_with_std}
    \begin{tabular}{c|c|c|c|c|c}
    \toprule
        \multicolumn{2}{c|}{Model} & \multicolumn{2}{c|}{MTST} & \multicolumn{2}{c}{PatchTST}  \\ \midrule
        Dataset & T & MSE & MAE & MSE & MAE \\ \midrule
        \multirow{4}{*}{Traffic} & 96 & 0.355 $\pm$ 0.0005 & 0.243 $\pm$ 0.0009 & 0.367  $\pm$  0.0006 & 0.250  $\pm$  0.0007 \\ 
         & 192 & 0.375 $\pm$ 0.0008 & 0.252 $\pm$ 0.0005 & 0.386  $\pm$  0.0004 & 0.259  $\pm$  0.0004 \\ 
         & 336 & 0.387 $\pm$ 0.0005 & 0.257 $\pm$ 0.0009 & 0.399  $\pm$  0.0010 & 0.267  $\pm$  0.0016 \\ 
         & 720 & 0.423 $\pm$ 0.0012 & 0.281 $\pm$ 0.0008 & 0.438  $\pm$  0.0097 & 0.291  $\pm$  0.0104 \\ \midrule
        \multirow{4}{*}{Electricity} & 96 & 0.127 $\pm$ 0.0003 & 0.221 $\pm$ 0.0006 & 0.130  $\pm$  0.0006 & 0.223  $\pm$  0.0006 \\ 
        & 192 & 0.145 $\pm$ 0.0000 & 0.238 $\pm$ 0.0003 & 0.148 $\pm $ 0.0002 & 0.240  $\pm$  0.0002 \\ 
        & 336 & 0.162 $\pm$ 0.0003 & 0.257 $\pm$ 0.0003 & 0.166  $\pm$  0.0006 & 0.260  $\pm$  0.0006 \\ 
        & 720 & 0.198 $\pm$ 0.0007 & 0.289 $\pm$ 0.0002 & 0.202  $\pm$  0.0006 & 0.292  $\pm$  0.0006 \\ \midrule
        \multirow{4}{*}{Weather} & 96 & 0.151 $\pm$ 0.0008 & 0.199 $\pm$ 0.0016 & 0.153  $\pm$  0.0024 & 0.200  $\pm$  0.0023 \\ 
        & 192 & 0.194 $\pm$ 0.0005 & 0.240 $\pm$ 0.0005 & 0.198  $\pm$  0.0015 & 0.243  $\pm$  0.0010 \\ 
        & 336 & 0.247 $\pm$ 0.0008 & 0.282 $\pm$ 0.0005 & 0.249  $\pm$  0.0012 & 0.284  $\pm$  0.0014 \\ 
        & 720 & 0.321 $\pm$ 0.0012 & 0.335 $\pm$ 0.0009 & 0.319  $\pm$  0.0002 & 0.335  $\pm$  0.0003 \\ \midrule
        \multirow{4}{*}{ETTh1} & 96 & 0.360 $\pm$ 0.0017 & 0.391 $\pm$ 0.0012 & 0.375  $\pm$  0.0008 & 0.400  $\pm$  0.0004 \\ 
        & 192 & 0.397 $\pm$ 0.0022 & 0.415 $\pm$ 0.0009 & 0.413  $\pm$  0.0012 & 0.421  $\pm$  0.0006 \\ 
        & 336 & 0.391 $\pm$ 0.0016 & 0.419 $\pm$ 0.0009 & 0.428  $\pm$  0.0033 & 0.433  $\pm$  0.0028 \\ 
        & 720 & 0.433 $\pm$ 0.0021 & 0.460 $\pm$ 0.0021 & 0.446  $\pm$  0.0035 & 0.464  $\pm$  0.0027 \\ \midrule
        \multirow{4}{*}{ETTh2} & 96 & 0.257 $\pm$ 0.0000 & 0.325 $\pm$ 0.0000 & 0.275  $\pm$  0.0005 & 0.336  $\pm$  0.0006 \\ 
        & 192 & 0.310 $\pm$ 0.0005 & 0.361 $\pm$ 0.0009 & 0.339  $\pm$  0.0010 & 0.379  $\pm$  0.0014 \\ 
        & 336 & 0.303 $\pm$ 0.0008 & 0.367 $\pm$ 0.0008 & 0.329  $\pm$  0.0010 & 0.382  $\pm$  0.0027 \\ 
        & 720 & 0.372 $\pm$ 0.0005 & 0.417 $\pm$ 0.0008 & 0.378  $\pm$  0.0010 & 0.421  $\pm$  0.0009 \\ \midrule
        \multirow{4}{*}{ETTm1} & 96 & 0.287 $\pm$ 0.0009 & 0.340 $\pm$ 0.0005 & 0.289 $\pm $ 0.0009 & 0.342  $\pm$  0.0007 \\ 
        & 192 & 0.331 $\pm$ 0.0025 & 0.368 $\pm$ 0.0021 & 0332  $\pm$  0.0008 & 0.370  $\pm$  0.0007 \\ 
        & 336 & 0.364 $\pm$ 0.0014 & 0.391 $\pm$ 0.0008 & 0.366  $\pm$  0.0022  & 0.391  $\pm$  0.0012 \\ 
        & 720 & 0.416 $\pm$ 0.0017 & 0.422 $\pm$ 0.0008 & 0.420  $\pm$  0.0056  & 0.424  $\pm$  0.0033 \\ \midrule
        \multirow{4}{*}{ETTm2} & 96 & 0.162 $\pm$ 0.0000 & 0.251 $\pm$ 0.0005 & 0.164  $\pm$  0.0011  & 0.254  $\pm$  0.0010 \\ 
        & 192 & 0.220 $\pm$ 0.0005 & 0.291 $\pm$ 0.0005 & 0.222  $\pm$  0.0008 & 0.294  $\pm$  0.0014 \\ 
        & 336 & 0.272 $\pm$ 0.0005 & 0.326 $\pm$ 0.0005 & 0.278  $\pm$  0.0020 & 0.330  $\pm$  0.0010 \\ 
        & 720 & 0.358 $\pm$ 0.0000 & 0.380 $\pm$ 0.0005 & 0.365  $\pm$  0.0024 & 0.383  $\pm$  0.0010 \\ \bottomrule
    \end{tabular}
\end{table}

\begin{table*}[h!]
\centering
\tiny
\caption{Multivariate long-term forecasting results with MTST. Look back window $L{=}96$ and prediction lengths $T{\in}\{96, 192, 336, 720\} $.
The best
results are in {bold} and the second best are {\ul underlined}.
The baseline results are from the original papers except for DLinear, which is reproduced for $L{=}96$.}
\label{tab:L96}
\useunder{\uline}{\ul}{}
\setlength{\tabcolsep}{2.1pt}
\begin{tabular}{cc|cc|cc|cc|cc|cc|cc|cc|cc|cc|cc}
\toprule
\multicolumn{2}{c|}{Models}                             & \multicolumn{2}{c|}{MTST}                          & \multicolumn{2}{c|}{PatchTST}                      & \multicolumn{2}{c|}{Dlinear}                       & \multicolumn{2}{c|}{MICN}                          & \multicolumn{2}{c|}{TimesNet}                      & \multicolumn{2}{c|}{Crossformer}                   & \multicolumn{2}{c|}{Scaleformer}                   & \multicolumn{2}{c|}{Fedformer}                     & \multicolumn{2}{c|}{Autoformer}                    & \multicolumn{2}{c}{Pyformer}                      \\ \midrule
\multicolumn{2}{c|}{Metric}                             & \multicolumn{1}{c}{MSE} & \multicolumn{1}{c|}{MAE} & \multicolumn{1}{c}{MSE} & \multicolumn{1}{c|}{MAE} & \multicolumn{1}{c}{MSE} & \multicolumn{1}{c|}{MAE} & \multicolumn{1}{c}{MSE} & \multicolumn{1}{c|}{MAE} & \multicolumn{1}{c}{MSE} & \multicolumn{1}{c|}{MAE} & \multicolumn{1}{c}{MSE} & \multicolumn{1}{c|}{MAE} & \multicolumn{1}{c}{MSE} & \multicolumn{1}{c|}{MAE} & \multicolumn{1}{c}{MSE} & \multicolumn{1}{c|}{MAE} & \multicolumn{1}{c}{MSE} & \multicolumn{1}{c|}{MAE} & \multicolumn{1}{c}{MSE} & \multicolumn{1}{c}{MAE} \\ \midrule
\multicolumn{1}{c|}{\multirow{4}{*}{Traffic}}     & 96  & \textbf{0.422}          & \textbf{0.271}           & {\ul 0.477}             & {\ul 0.305}              & 0.650                   & 0.397                    & 0.519                   & 0.309                    & 0.593                   & 0.321                    & -                       & -                        & 0.564                   & 0.351                    & 0.587                   & 0.366                    & 0.613                   & 0.388                    & 2.085                   & 0.468                   \\
\multicolumn{1}{c|}{}                             & 192 & \textbf{0.437}          & \textbf{0.281}           & {\ul 0.471}             & {\ul 0.299}              & 0.600                   & 0.372                    & 0.537                   & 0.315                    & 0.617                   & 0.336                    & -                       & -                        & 0.570                   & 0.349                    & 0.604                   & 0.373                    & 0.616                   & 0.382                    & 0.867                   & 0.467                   \\
\multicolumn{1}{c|}{}                             & 336 & \textbf{0.451}          & \textbf{0.285}           & {\ul 0.485}             & 0.305                    & 0.606                   & 0.374                    & 0.534                   & 0.313                    & 0.629                   & 0.336                    & 0.530                   & {\ul 0.300}              & 0.576                   & 0.349                    & 0.621                   & 0.383                    & 0.622                   & 0.337                    & 0.869                   & 0.469                   \\
\multicolumn{1}{c|}{}                             & 720 & \textbf{0.490}          & \textbf{0.309}           & {\ul 0.518}             & 0.325                    & 0.646                   & 0.395                    & 0.577                   & 0.325                    & 0.640                   & 0.350                    & 0.573                   & {\ul 0.313}              & 0.602                   & 0.360                    & 0.626                   & 0.382                    & 0.660                   & 0.408                    & 0.881                   & 0.473                   \\ \midrule
    \multicolumn{1}{c|}{\multirow{4}{*}{Electricity}} & 96  & \textbf{0.160}          & \textbf{0.248}           & 0.174                   & {\ul 0.259}              & 0.194                   & 0.277                    & {\ul 0.164}       & 0.269                    & 0.168            & 0.272                    & -                       & -                        & 0.182                   & 0.297                    & 0.193                   & 0.308                    & 0.201                   & 0.317                    & 0.386                   & 0.449                   \\
\multicolumn{1}{c|}{}                             & 192 & \textbf{0.171}          & \textbf{0.263}           & 0.178                   & {\ul 0.265}              & 0.193                   & 0.280                    & {\ul 0.177}             & 0.285                    & 0.184                   & 0.289                    & -                       & -                        & 0.188                   & 0.300                    & 0.201                   & 0.315                    & 0.222                   & 0.334                    & 0.386                   & 0.443                   \\
\multicolumn{1}{c|}{}                             & 336 & \textbf{0.188}          & \textbf{0.281}           & 0.196                   & {\ul 0.282}              & 0.207                   & 0.296                    & {\ul 0.193}             & 0.304                    & 0.198                   & 0.300                    & 0.404                   & 0.423                    & 0.210                   & 0.324                    & 0.214                   & 0.329                    & 0.231                   & 0.338                    & 0.378                   & 0.443                   \\
\multicolumn{1}{c|}{}                             & 720 & 0.230                   & \textbf{0.315}           & 0.237                   & {\ul 0.316}              & 0.242                   & 0.328                    & \textbf{0.212}          & 0.321                    & {\ul 0.220}             & 0.320                    & 0.433                   & 0.438                    & 0.232                   & 0.339                    & 0.246                   & 0.355                    & 0.254                   & 0.361                    & 0.376                   & 0.445                   \\ \midrule
\multicolumn{1}{c|}{\multirow{4}{*}{Weather}}     & 96  & 0.175                   & \textbf{0.216}           & 0.178                   & {\ul 0.219}              & 0.197                   & 0.255                    & \textbf{0.161}          & 0.229                    & 0.172                   & 0.220                    & -                       & -                        & {\ul 0.163}             & 0.226                    & 0.217                   & 0.296                    & 0.266                   & 0.336                    & 0.896                   & 0.556                   \\
\multicolumn{1}{c|}{}                             & 192 & \textbf{0.219}          & \textbf{0.255}           & 0.224                   & {\ul 0.259}              & 0.239                   & 0.297                    & {\ul 0.220}             & 0.281                    & \textbf{0.219}          & 0.261                    & -                       & -                        & 0.221                   & 0.290                    & 0.276                   & 0.336                    & 0.307                   & 0.367                    & 0.622                   & 0.624                   \\
\multicolumn{1}{c|}{}                             & 336 & \textbf{0.276}          & \textbf{0.296}           & 0.278                   & {\ul 0.298}              & 0.284                   & 0.332                    & {\ul 0.278}             & 0.331                    & 0.280                   & 0.306                    & 0.495                   & 0.515                    & 0.282                   & 0.340                    & 0.339                   & 0.380                    & 0.359                   & 0.395                    & 0.739                   & 0.753                   \\
\multicolumn{1}{c|}{}                             & 720 & 0.351                   & \textbf{0.346}           & 0.350                   & \textbf{0.346}           & {\ul 0.348}             & 0.385                    & \textbf{0.311}          & {\ul 0.356}              & 0.365                   & 0.359                    & 0.526                   & 0.542                    & 0.369                   & 0.396                    & 0.403                   & 0.428                    & 0.419                   & 0.428                    & 1.004                   & 0.934                   \\ \midrule
\multicolumn{1}{c|}{\multirow{4}{*}{ETTh1}}       & 96  & \textbf{0.376}          & \textbf{0.393}           & {\ul 0.393}             & 0.408                    & 0.383                   & {\ul 0.396}              & 0.421                   & 0.431                    & 0.384                   & 0.402                    & -                       & -                        & -                       & -                        & \textbf{0.376}          & 0.419                    & 0.449                   & 0.459                    & 0.664                   & 0.612                   \\
\multicolumn{1}{c|}{}                             & 192 & 0.429                   & \textbf{0.422}           & 0.445                   & 0.434                    & 0.433                   & 0.426                    & 0.474                   & 0.487                    & 0.436                   & 0.429                    & -                       & -                        & -                       & -                        & \textbf{0.420}          & 0.448                    & 0.500                   & 0.482                    & 0.790                   & 0.681                   \\
\multicolumn{1}{c|}{}                             & 336 & {\ul 0.444}             & \textbf{0.436}           & 0.484                   & {\ul 0.451}              & 0.491                   & 0.467                    & 0.569                   & 0.551                    & 0.491                   & 0.469                    & \textbf{0.440}          & 0.461                    & -                       & -                        & 0.459                   & 0.465                    & 0.521                   & 0.496                    & 0.891                   & 0.738                   \\
\multicolumn{1}{c|}{}                             & 720 & \textbf{0.469}          & \textbf{0.466}           & {\ul 0.480}             & {\ul 0.471}              & 0.527                   & 0.518                    & 0.770                   & 0.672                    & 0.521                   & 0.500                    & 0.519                   & 0.524                    & -                       & -                        & 0.506                   & 0.507                    & 0.514                   & 0.512                    & 0.963                   & 0.782                   \\ \midrule
\multicolumn{1}{c|}{\multirow{4}{*}{ETTh2}}       & 96  & \textbf{0.276}          & \textbf{0.333}           & {\ul 0.294}             & {\ul 0.343}              & 0.329                   & 0.380                    & 0.299                   & 0.364                    & 0.340                   & 0.374                    & -                       & -                        & -                       & -                        & 0.346                   & 0.388                    & 0.358                   & 0.397                    & 0.645                   & 0.597                   \\
\multicolumn{1}{c|}{}                             & 192 & \textbf{0.353}          & \textbf{0.382}           & {\ul 0.377}             & {\ul 0.393}              & 0.431                   & 0.443                    & 0.441                   & 0.454                    & 0.402                   & 0.414                    & -                       & -                        & -                       & -                        & 0.429                   & 0.439                    & 0.456                   & 0.452                    & 0.788                   & 0.683                   \\
\multicolumn{1}{c|}{}                             & 336 & \textbf{0.357}          & \textbf{0.395}           & {\ul 0.381}             & {\ul 0.409}              & 0.455                   & 0.460                    & 0.654                   & 0.567                    & 0.452                   & 0.452                    & -                       & -                        & -                       & -                        & 0.496                   & 0.487                    & 0.482                   & 0.486                    & 0.907                   & 0.747                   \\
\multicolumn{1}{c|}{}                             & 720 & \textbf{0.406}          & \textbf{0.430}           & {\ul 0.412}             & {\ul 0.433}              & 0.773                   & 0.631                    & 0.956                   & 0.716                    & 0.462                   & 0.468                    & -                       & -                        & -                       & -                        & 0.463                   & 0.474                    & 0.515                   & 0.511                    & 0.963                   & 0.783                   \\ \midrule
\multicolumn{1}{c|}{\multirow{4}{*}{ETTm1}}       & 96  & 0.323                   & \textbf{0.360}           & {\ul 0.321}             & \textbf{0.360}           & 0.346                   & 0.374                    & \textbf{0.316}          & {\ul 0.362}              & 0.338                   & 0.375                    & -                       & -                        & -                       & -                        & 0.379                   & 0.419                    & 0.505                   & 0.475                    & 0.543                   & 0.510                   \\
\multicolumn{1}{c|}{}                             & 192 & {\ul 0.363}             & {\ul 0.386}              & \textbf{0.362}          & \textbf{0.384}           & 0.383                   & 0.393                    & 0.363                   & 0.390                    & 0.374                   & 0.387                    & -                       & -                        & -                       & -                        & 0.426                   & 0.441                    & 0.553                   & 0.496                    & 0.557                   & 0.537                   \\
\multicolumn{1}{c|}{}                             & 336 & {\ul 0.393}             & {\ul 0.406}              & \textbf{0.392}          & \textbf{0.402}           & 0.417                   & 0.418                    & 0.408                   & 0.426                    & 0.410                   & 0.411                    & 0.404                   & 0.427                    & -                       & -                        & 0.445                   & 0.459                    & 0.621                   & 0.537                    & 0.754                   & 0.655                   \\
\multicolumn{1}{c|}{}                             & 720 & {\ul 0.453}             & {\ul 0.441}              & \textbf{0.450}          & \textbf{0.435}           & 0.479                   & 0.457                    & 0.481                   & 0.476                    & 0.478                   & 0.450                    & 0.569                   & 0.528                    & -                       & -                        & 0.543                   & 0.490                    & 0.671                   & 0.561                    & 0.908                   & 0.724                   \\ \midrule
\multicolumn{1}{c|}{\multirow{4}{*}{ETTm2}}       & 96  & \textbf{0.174}          & \textbf{0.256}           & {\ul 0.178}             & {\ul 0.260}              & 0.187                   & 0.281                    & 0.179                   & 0.275                    & 0.187                   & 0.267                    & -                       & -                        & -                       & -                        & 0.203                   & 0.287                    & 0.255                   & 0.339                    & 0.435                   & 0.507                   \\
\multicolumn{1}{c|}{}                             & 192 & \textbf{0.243}          & \textbf{0.302}           & {\ul 0.249}             & {\ul 0.307}              & 0.272                   & 0.349                    & 0.307                   & 0.376                    & 0.249                   & 0.309                    & -                       & -                        & -                       & -                        & 0.269                   & 0.328                    & 0.281                   & 0.340                    & 0.730                   & 0.673                   \\
\multicolumn{1}{c|}{}                             & 336 & \textbf{0.301}          & \textbf{0.340}           & {\ul 0.313}             & {\ul 0.346}              & 0.344                   & 0.395                    & 0.325                   & 0.388                    & 0.321                   & 0.351                    & -                       & -                        & -                       & -                        & 0.325                   & 0.366                    & 0.339                   & 0.372                    & 1.201                   & 0.845                   \\
\multicolumn{1}{c|}{}                             & 720 & \textbf{0.397}          & \textbf{0.395}           & {\ul 0.400}             & {\ul 0.398}              & 0.438                   & 0.444                    & 0.502                   & 0.490                    & 0.408                   & 0.403                    & -                       & -                        & -                       & -                        & 0.421                   & 0.415                    & 0.433                   & 0.432                    & 3.625                   & 1.451                   \\ \midrule
\multicolumn{2}{c|}{\# Rank 1}                          & 19                      & 25                       & 3                       & 5                        & 0                       & 0                        & 4                       & 0                        & 1                       & 0                        & 1                       & 0                        & 0                       & 0                        & 2                       & 0                        & 0                       & 0                        & 0                       & 0    
\\ \bottomrule
\end{tabular}
\end{table*}

\section{ADDITIONAL EXPERIMENTAL RESULTS}
\subsection{Forecasting Results with Standard Deviations}
To quantify the variability of our experimental results in Table~\ref{tab:L336}, we repeat each experiment three times with different random seeds (2021, 2022, and 2023), resulting in different initializations of the model parameters in each run. The results shown in Table~\ref{tab:L336} of the main paper is obtained from the trained model, which was initialized by setting the random seed to 2021.
The mean and standard derivation of the MSE of MAE across multiple trials are reported in Table~\ref{tab:result_with_std}. 
The variances are small, which indicates that using a different random seed has minimal impact on the forecast.

% $\bp_{ij}{:=}\func{sign}(i-j) p(|i-j|)$,
% where $\func{sign}$ denotes the sign function and the function $\func{PE}: \mathbb{Z}^+{\rightarrow}\sR^{D_{pos}}$ is defined as
% \begin{align}
% \func{PE}_{2t}(i)&{:=} \sin(i/10000^{2t/D_{pos}})\,,\nonumber\\
% \func{PE}_{2t{+}1}(i)&{:=} \cos(i/10000^{2t/D_{pos}})\,,\label{eq:pe_design_apx}
% \end{align}
% for $t \in \{1, \ldots, D_{pos}/2\}$.

\subsection{Results for a Shorter Look-back Window (\texorpdfstring{$L{=}96$}{L=96})}
For a fair comparison, we also conduct the experiments under the $L{=}96$ setting that is the default for MICN, TimesNet, and other transformer-based models, except PatchTST. For the shorter look-back window, we slightly modify the patch length for each branch since we need $P{<}L$ to obtain the tokens.
By default, MTST has 2 layers and 2 branches across all datasets and horizons. At each layer, the patch length is set to $P{=}[12,16]$. Other hyper-parameters follows the $L=336$ setting in Table~\ref{tab:L336}.

As shown in Table~\ref{tab:L96}, 
% MTST emerges as the leading performer, showcasing its robustness and accuracy in forecasting. 
MTST secures Rank 1 in 19 cases out of 28 cases in terms of MSE and 25 cases in terms of MAE.
% We also rank the algorithms in Table 2 based on their MSE and order them based on their average rank across seven datasets and four prediction horizons. Figure.~7 shows the boxplot of rank. We observe that the proposed MTST achieves the best average rank and lowest variability across all settings.This remarkable feat underscores the model's prowess in capturing intricate temporal dependencies within a shorter look-back window. 
As a strong contender, PatchTST claims top performance on ETTm1 dataset. This matches our observations in the ablation study of the usefulness of multi-resolution, which illustrates that the one branch patch-based TST could have slightly better performance for small datasets because of the overfitting on high-resolution components. 
% Additionally, MICN  displayed strength in 4 MSE scenarios and demonstrated exceptional performance with zero MAE rankings, reinforcing its proficiency in minimizing larger errors. 
MTST's consistent success across various metrics and look-back windows highlights its adaptability and reliability in addressing time-series forecasting challenges, making it a compelling choice for real-world applications.

\subsection{Results on ILI Dataset}
The influenza-like illness (ILI) dataset\footnote{https://gis.cdc.gov/grasp/fluview/fluportaldashboard.html} contains the weekly time-series of ratio of patients seen with ILI and the total number of the patients in the United States between 2002 and 2021.
It has 7 variates and 966 timestamps, and is thus a smaller dataset compared to the others we study. Therefore, we use a different setting, following the experimental setup in~\citet{nie2023TimeSeriesWorth}, so that the prediction horizon $T{\in}\{24, 36, 48, 60\}$ and the look-back window $L{=}104$.
We report the baseline results from \citet{nie2023TimeSeriesWorth} except for TimesNet and MICN, for which we reproduce 
their results with $L{=}104$ based on the officially released code. Note that the results for Fedformer and Autoformer are the best from multiple look-back windows $L{\in}\{24, 36, 48, 60, 104, 144\}$.

For the hyperparameters for this dataset, we have $N{=}1$ layer with $B{=}2$ branches. At each branch, the transformer consists of an attention mechanism with 4 heads, and the 2-layer feed-forward
network with hidden dimension of 128 and output dimension of 32. The patch length for each branch $P = [12,24]$ and stride length $S = [2,4]$. The findings presented in Table~\ref{table:ili336} indicate that MTST surpasses all baseline models across all horizons in terms of the MSE metric.

\begin{table*}[ht]
\centering
\scriptsize
\caption{Multivariate long-term forecasting results for the ILI dataset with $L{=}104$ and $T{\in}\{24, 36, 48, 60\}$.
\textbf{Bold} and \ul{underlined} denote the best and second-best results respectively.
$^*$ indicates statistically significant difference between the top-2 results.}
\useunder{\uline}{\ul}{}
\setlength{\tabcolsep}{5pt}
\begin{tabular}{cc|ll|cc|cc|cc|cc|cc|cc}
\toprule
\multicolumn{2}{c|}{Models}                             & \multicolumn{2}{c|}{MTST}       & \multicolumn{2}{c|}{PatchTST} & \multicolumn{2}{c|}{DLinear} & \multicolumn{2}{c|}{MICN}      & \multicolumn{2}{c|}{TimesNet}    & \multicolumn{2}{c|}{Fedformer}   & \multicolumn{2}{c}{Autoformer}  \\ \midrule
\multicolumn{1}{c|}{Dataset} & $T$                            & \multicolumn{1}{c}{MSE}           & \multicolumn{1}{c|}{MAE}            & MSE         & MAE         & MSE         & MAE            & MSE           & MAE            & MSE           & MAE             & MSE         & MAE            & MSE         & MAE                          \\ \midrule

\multicolumn{1}{c|}{\multirow{4}{*}{ILI}}       & 24   & \textbf{1.499*} & \textbf{0.790*} & {\ul 1.522}   & {\ul 0.814}          & 2.215             & 1.081              &      2.345       &    1.043         &     2.157         &     0.978   & 2.624 & 1.095& 2.906& 1.182 \\
\multicolumn{1}{c|}{}                             & 36 & \textbf{1.413*} & \textbf{0.789*} & {\ul 1.430}   & {\ul 0.834}         &1.963 & 0.963  &        2.330      &     1.001        &        2.318       &    1.031      &  2.516 & 1.021 & 2.585 & 1.038       \\
\multicolumn{1}{c|}{}                             & 48 & \textbf{1.605*} & {\ul 0.877} & {\ul 1.673}   & \textbf{0.854*}   & 2.130         & 1.024        &   2.386          &     1.051        &     2.121          &   1.005   & 2.505 & 1.041 & 3.024 & 1.145  \\
\multicolumn{1}{c|}{}                             & 60 & \textbf{1.499*} & \textbf{0.814*} & {\ul 1.529}   & {\ul 0.862}    & 2.368         & 1.096        &    2.616         &    1.131         &       1.975        &      0.975      &  2.742 & 1.122 & 2.761 & 1.114    \\ 

% \multicolumn{2}{c|}{\#Rank-1st (total=28)}        & \multicolumn{1}{c}{28} & \multicolumn{1}{c|}{26} &  1         &  0   & 0            & 3               &  0               & 0 & 0                      & 0 & 0 & 0 &  0 & 0        &  0 & 0         \\ 
\bottomrule
\end{tabular}
\label{table:ili336}
\end{table*}

\section{RELATED WORK}
\label{sec:rel_apx}

\subsection{Long-term Time-Series Forecasting}
% \paragraph{Early deep learning models}
Multivariate time-series forecasting has been an essential research focus for decades, 
being developed from various conventional statistical models~\cite{makridakis1997ArmaModelsBox,hyndman2008ForecastingExponentialSmoothing,holt2004ForecastingSeasonalsTrends}
to diverse deep-learning techniques.
% such as ARIMA~\cite{}. 
% Inspired by the success of deep learning, several
% methods began to approach long-term time series forecasting with deep learning networks.  
% In recent years, various neural network architectures for sequence modeling and time series forecasting have been introduced.
% In the wave of deep learning, 
% various 
Viewing time-series as sequential data and inspired by early auto-regressive statistical models,
earlier deep learning methods propose to model time-series with  
recurrent neural networks (RNNs)~\cite{salinas2020DeeparProbabilisticForecasting, smyl2020hybrid, lai2018ModelingLongShort, lim2021TemporalFusionTransformers, wen2017Multi-horizonQuantile}.
Besides RNNs, convolution neural networks~(CNNs) have also been widely used to extract features from time-series~\cite{bai2018EmpiricalEvaluationGeneric, wang2023MicnMultiscaleLocal, cui2016MultiScaleConvolution}
% \mlh{yitian add more citations}.
However, most RNN and CNN-based approaches struggle to capture long-term dependency, which is crucial for long-term time-series forecasting.
Recently, several works have proposed model architectures based on multiple-layer perceptrons (MLPs)~\cite{oreshkin2021NbeatsNeuralNetwork,challu2022NHiTSNeuralHierarchical,vijay2023TsmixerLightweightMLP} and demonstrated satisfactory performance for long-term forecasting.

Driven by the same motivation, numerous transformer-based models have been introduced to the time-series domain to better capture long-range dependency.
Due to the high computational cost of attention mechanisms, earlier works in time-series transformers (TSTs) focus on the efficiency of learning long time-series sequences~\cite{li2019EnhancingLocalityBreaking, zhou2021InformerEfficientTransformer}.
Another line of work, concentrating on prediction performance, incorporates various inductive biases of time-series into achitecture designs:
trend-seasonal decomposition and auto-correlation in Autoformer~\cite{wu2021AutoformerDecompositionTransformers}, multi-scale feature representation in Pyraformer ~\cite{liu2021PyraformerLow-complexityPyramidal} and ScaleFormer~\cite{shabani2023Scaleformer}
as well as signal patterns of multiple frequencies in FEDformer~\cite{zhou2022FEDformerFrequencyEnhanced}.

The aforementioned TSTs are all based on timestamp-level tokenization. A recent study argued that these TSTs are incapable of learning temporal patterns and can be outperformed by a simple linear model~\cite{zeng2023AreTransformersEffective}. 
Subsequently,  PatchTST~\cite{nie2023TimeSeriesWorth}  introduced patch-level tokenization~\cite{dosovitskiy2020ImageWorth16x16} and reached \textit{state-of-the-art} performance for various time-series benchmarks.
Another related approach is   CrossFormer~\cite{zhang2023CrossformerTransformerUtilizing}, which aims to capture both cross-time and cross-variate dependencies using attention.
However, due to the non-cognizance of the original chronological order from patching, 
how to incorporate the inductive biases of time-series into TSTs with patch-level tokenization remains an open question.
According to the best of our knowledge, our proposed method is one of the first to incorporate multi-scale features as an inductive bias into patch-based time-series transformers.

% LogTrans~\cite{li2019EnhancingLocalityBreaking} utilizes convolutional self-attention layers with a LogSparse design to capture local information with favorable memory requirements.

% Informer~\cite{zhou2021InformerEfficientTransformer} introduces a ProbSparse self-attention mechanism incorporating distilling techniques to efficiently extract crucial keys.

% Autoformer~\cite{wu2021AutoformerDecompositionTransformers} incorporates concepts from traditional time series analysis methods such as decomposition and auto-correlation. FEDformer~\cite{zhou2022FEDformerFrequencyEnhanced} adopts a Fourier enhanced structure, ensuring linear complexity. Additionally, Pyraformer \cite{liu2021PyraformerLow-complexityPyramidal} employs a pyramidal attention module featuring inter-scale and intra-scale connections, achieving linear complexity as well.

% Most of these architectures form tokens using a single time-step, which limits their usefulness in learning temporal patterns. Although PatchTST and CrossFormer overcome this issue by utilizing fixed-length patches as tokens, they still lack flexibility in terms of their capability to learn seasonalities at varying scales.  

% List: LogTrans, Informer, Autoformer, FedFormer, PyraFormer,  Scaleformer, Crossformer, DSformer, PreFormer

\subsection{Multi-scale Feature Learning}

Combining multiple scales of contextual features has proven beneficial in processing complex, information-rich signals such as images, videos, and time-series.
% , dense prediction problems, 
% including but not limited to time-series forecasting
% and image semantic segmentation~\cite{sercu2016DensePredictionSequences}.
% which drives the development of multi-scale learning methods.
For example, several works proposed learning of multi-scale features in the image domain, including
learning with convolution of different scales~\cite{yu2016MultiScaleContextAggregation, chen2017DeepLabSemanticImage, szegedy2016RethinkingInceptionArchitecture}, and constructing multi-resolution representations via subsampling~\cite{ronneberger2015UNetConvolutionalNetworks, lin2017FeaturePyramidNetworks}.
Notably, following the introduction of patch-based transformers in computer vision, several works have constructed hierarchical representations via merging patches instead of subsampling~\cite{wang2021PyramidVisionTransformer,liu2021SwinTransformerHierarchical, bolya2023TokenMergingYour}.

Similar techniques have been introduced for modeling time-series in recent years.
For non-transformer-based models,
several works propose to capture multi-scale features with different operators in a multi-branch architecture:
MCNN~\cite{cui2016MultiScaleConvolution} utilizes different downsampling transformations followed by convolution and max-pooling;
MICN~\cite{wang2023MicnMultiscaleLocal} incorporates convolution with different kernel sizes in each branch;
TAMS-RNNs~\cite{chen2021TimeAwareMulti} propose multi-scale RNNs with multiple hidden states. 
NHits~\cite{challu2022NHiTSNeuralHierarchical} introduces multi-rate signal sampling schemes to the NBeats architecture to model multi-granularity features.

Among transformer-based models,
Pyraformer~\cite{liu2021PyraformerLow-complexityPyramidal} proposes pyramidal attention with inter-scale and intra-scale connections to build multi-resolution representations with a focus on reducing computational complexity.
Scaleformer~\cite{shabani2023Scaleformer} uses mean-pooling for downsampling to reach multi-resolution representations for timestamp-level tokens.
However, pooling techniques typically result in sub-optimal representation when applied to patch-level tokens, due to non-cognizance of the original chronological order~\cite{marin2023TokenPoolingVision}. 
Therefore, we explore the alternative method to construct multi-resolution representations with adjusting patch sizes.

\section{FORECAST VISUALIZATION}
\label{sec:app:visual}
\begin{figure*}[htbp]
\centering
    \begin{minipage}{0.85\textwidth}
    \centering
        \includegraphics[width=\textwidth]{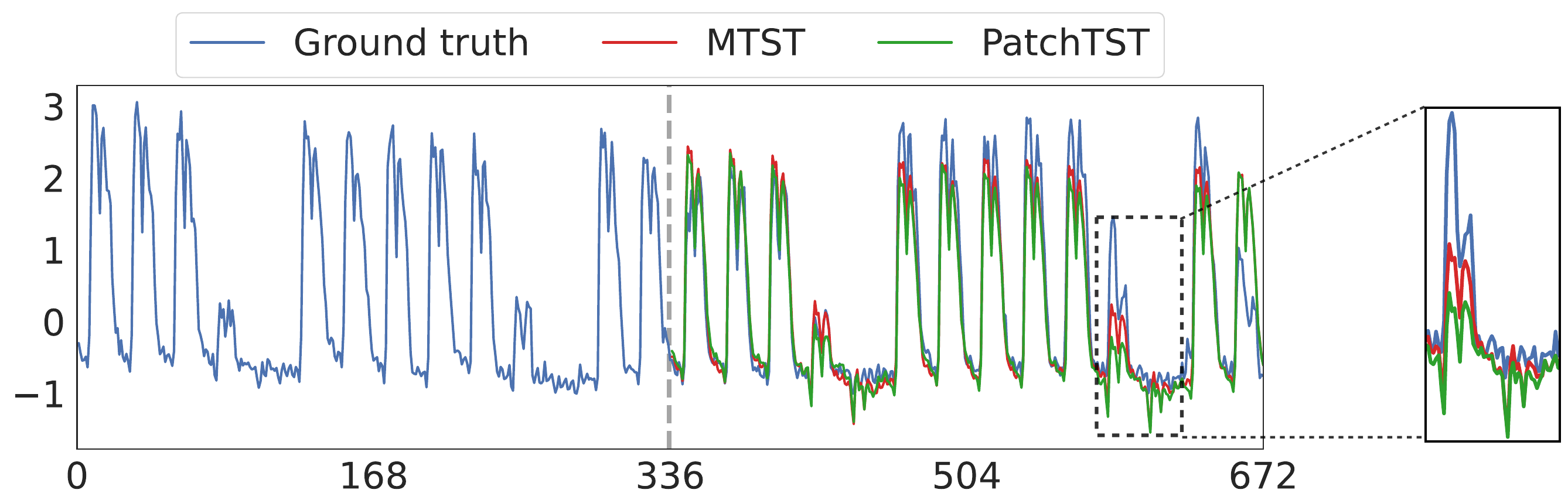}
        % \subcaption{$310$-th variate on $4050$-th example on \text{Electricity}.}
        \subcaption{Comparison of MTST and PatchTST}
        \label{fig:visual_electricity1_apx}
    \end{minipage} \hfill
    \begin{minipage}{0.85\textwidth}
        \centering
        \includegraphics[width=\textwidth]{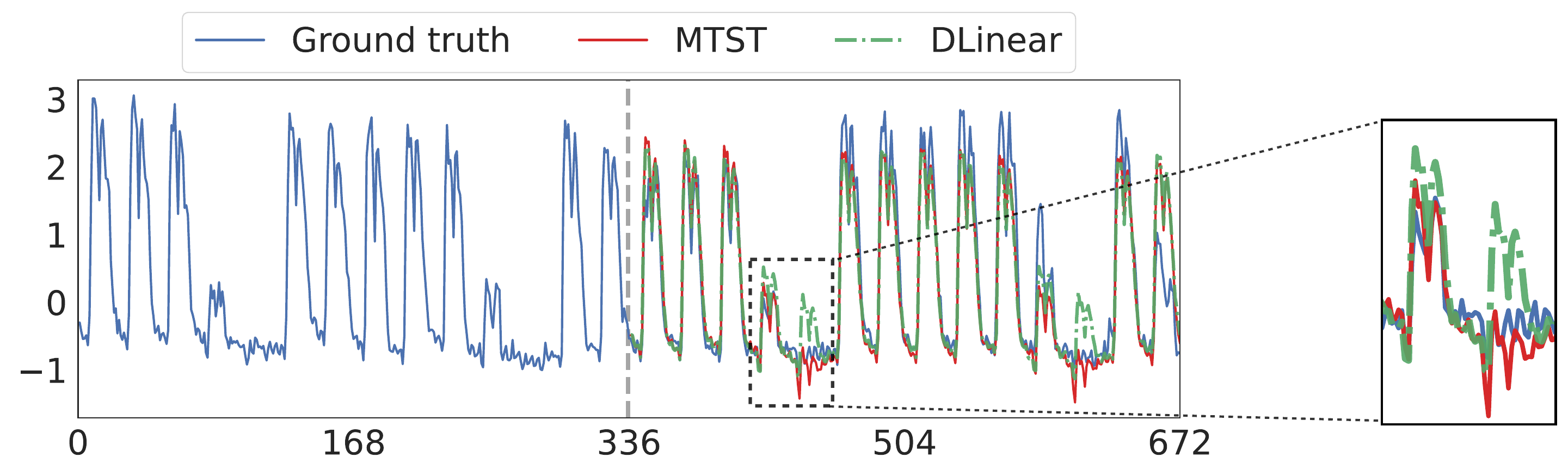}
        \subcaption{Comparison of MTST and DLinear}
        \label{fig:visual_electricity2_apx}
    \end{minipage}
       \begin{minipage}{0.85\textwidth}
    \centering
        \includegraphics[width=\textwidth]{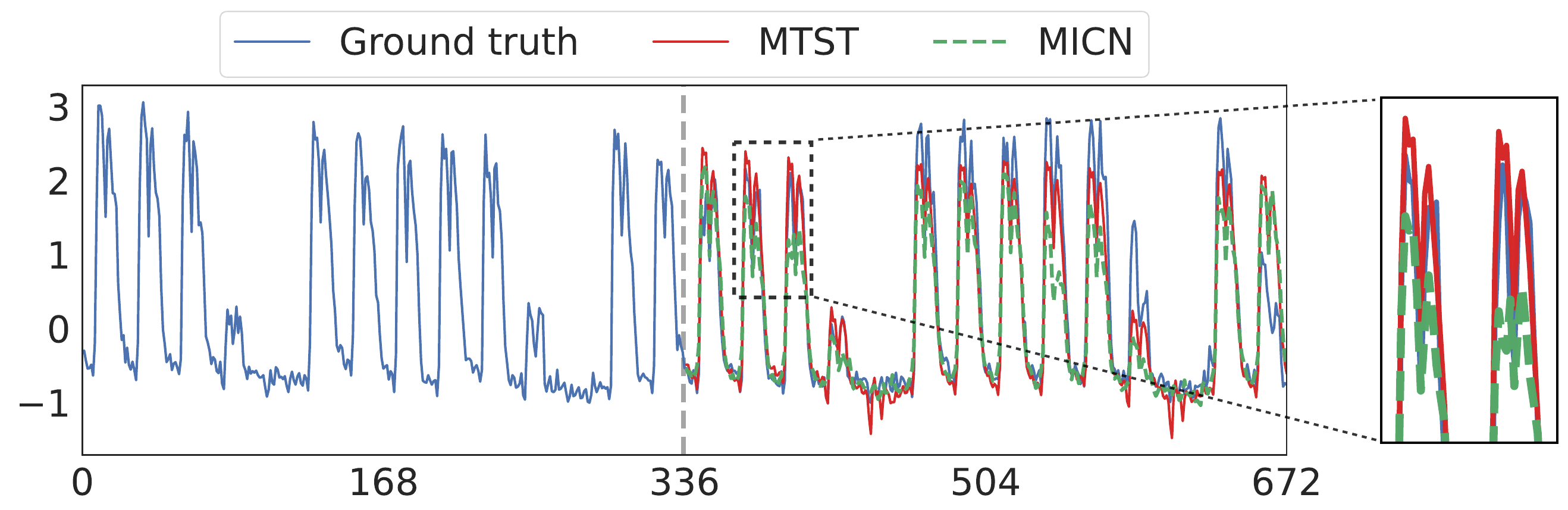}
        % \subcaption{$310$-th variate on $4050$-th example on \text{Electricity}.}
        \subcaption{Comparison of MTST and MICN.}
        \label{fig:visual_electricity3_apx}
    \end{minipage} \hfill
    \caption{$11$-th variate on $28$-th test sample on \text{Electricity}.}
    \label{fig:visual_electricity_apx}
\end{figure*}

\begin{figure*}[htbp]
\centering
    \begin{minipage}{0.85\textwidth}
    \centering
        \includegraphics[width=\textwidth]{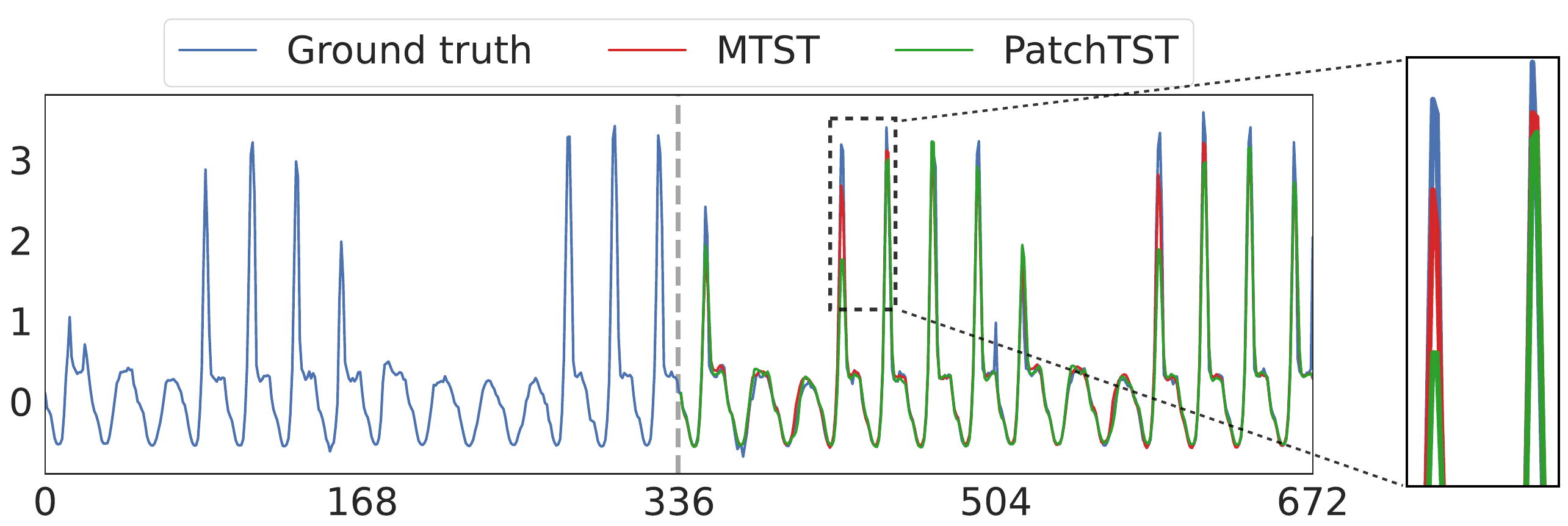}
        % \subcaption{$310$-th variate on $4050$-th example on \text{Electricity}.}
        \subcaption{Comparison of MTST and PatchTST}
        \label{fig:visual_traffic1_apx}
    \end{minipage} \hfill
    \begin{minipage}{0.85\textwidth}
        \centering
        \includegraphics[width=\textwidth]{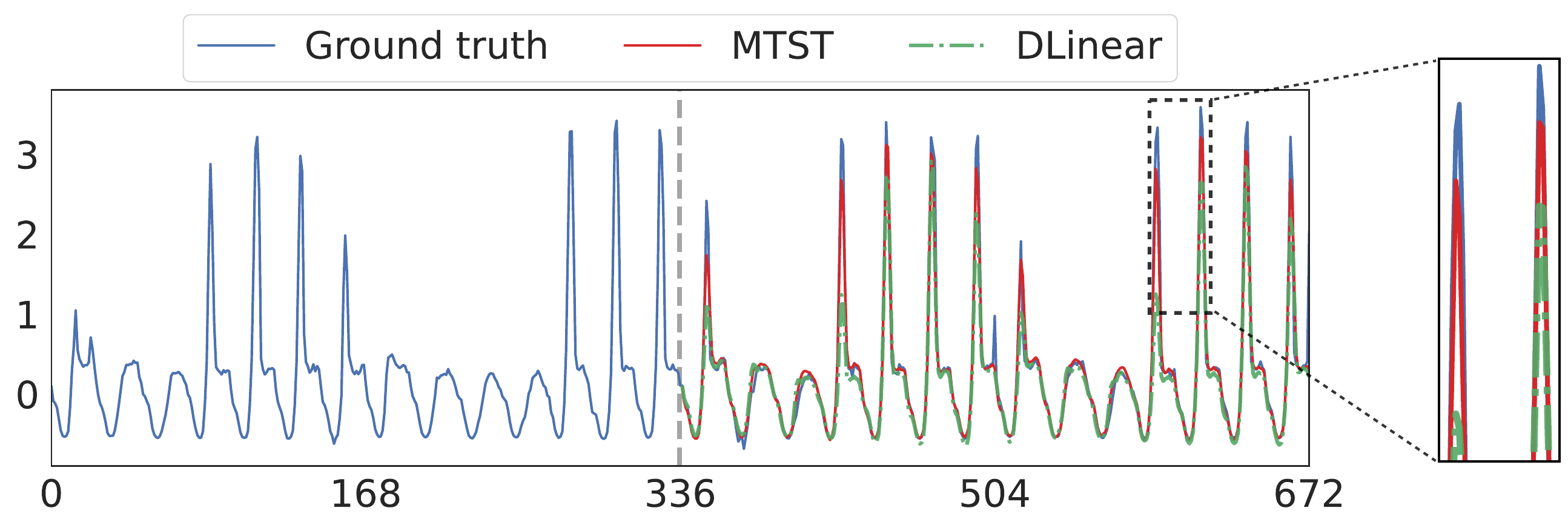}
        \subcaption{Comparison of MTST and DLinear}
        \label{fig:visual_traffic2_apx}
    \end{minipage}
       \begin{minipage}{0.85\textwidth}
    \centering
        \includegraphics[width=\textwidth]{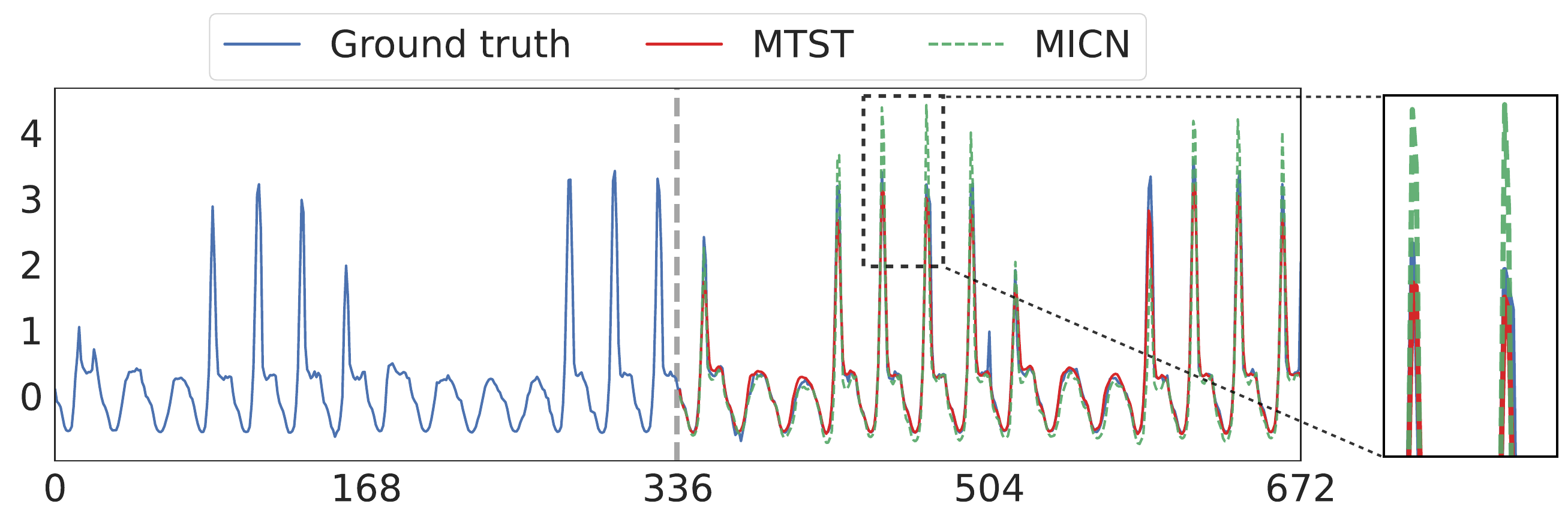}
        % \subcaption{$310$-th variate on $4050$-th example on \text{Electricity}.}
        \subcaption{Comparison of MTST and MICN.}
        \label{fig:visual_traffic3_apx}
    \end{minipage} \hfill
    \caption{$11$-th variate on $765$-th test sample on \text{Traffic}.}
    \label{fig:visual_traffic_apx}
\end{figure*}

\begin{figure*}[htbp]
\centering
    \includegraphics[width=0.9\textwidth]{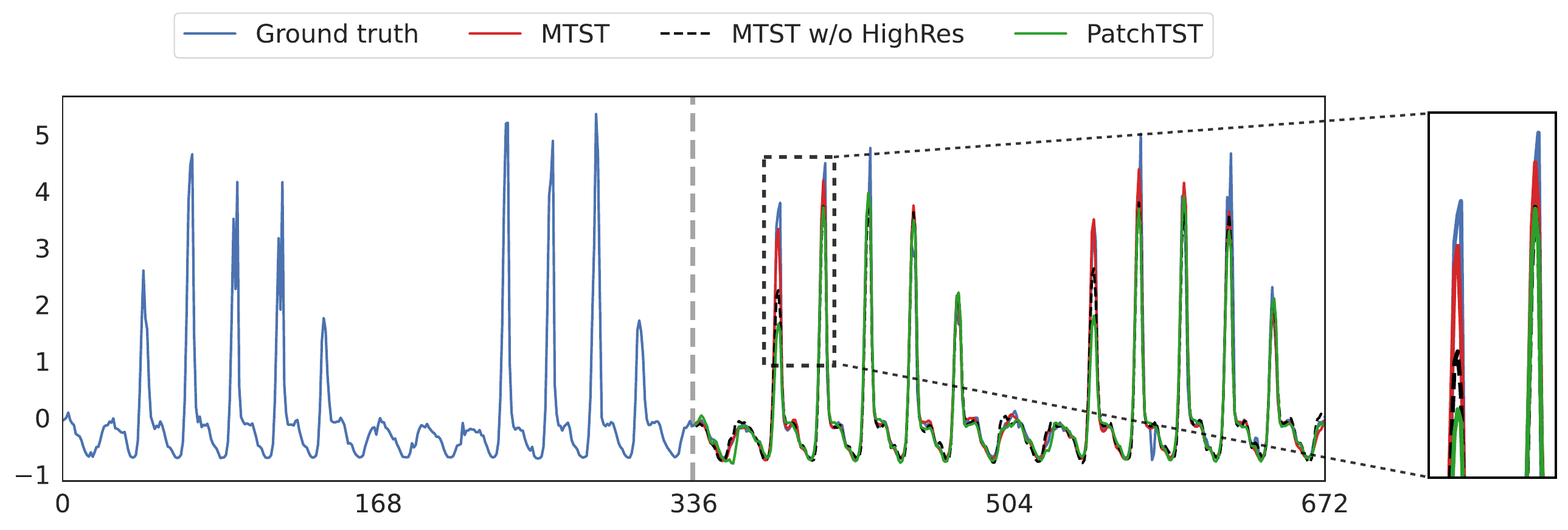}
\caption{$4$-th variate on $804$-th test sample on \text{Traffic}.}
\label{fig:visual_example_apx}
\end{figure*}

 We visualize the long-term forecasting results of MTST and other baselines in Figure~\ref{fig:visual_electricity_apx} and Figure~\ref{fig:visual_traffic_apx}. Here we show the results of predicting 336 steps ahead on Electricity and Traffic dataset. In comparison to PatchTST and DLinear, as shown in Figure~\ref{fig:visual_electricity1_apx} and \ref{fig:visual_electricity2_apx}, MTST exhibits superior performance in capturing the weekend patterns within the data specifically. This proficiency is attributed to the low-resolution component, which captures long-term seasonality and trends. Moreover, from Figure~\ref{fig:visual_electricity3_apx} and Figure~\ref{fig:visual_traffic3_apx}, we can see that MTST outperforms in scale prediction compared to MICN. Furthermore, MTST demonstrates superior prediction accuracy, particularly when it comes to forecasting peaks in the Traffic dataset (Figure~\ref{fig:visual_traffic_apx}). This enhanced performance can potentially be attributed to the finer predictions generated by the high-resolution branch of MTST. To validate this hypothesis, we conducted an experiment where the high-resolution branch was removed. The results, as illustrated in Figure~\ref{fig:visual_example_apx}, clearly indicate a noticeable decline in performance of MTST w/o HighRes, emphasizing the significant contribution of the high-resolution branch in enabling finer predictions.
 
%  This capability highlights MTST's effectiveness of capturing both the short-term peaks characterized by high fluctuations, as well as the longer-term weekend patterns. 

% Meanwhile, Figure~4 provides the same example depicted in Figure 6b of the main paper. In this comparison, the impact of the low amplitude of the first high peak within the second weekly cycle in the look-back window is examined. PatchTST appears to be more affected by this low amplitude, whereas MTST demonstrates superior forecasting accuracy regarding the amplitude of the first high peak. This enhanced performance can potentially be attributed to the finer predictions generated by the high-resolution branch of MTST. To validate this hypothesis, we conducted an experiment where the high-resolution branch was removed. The results, as illustrated in Figure~2, clearly indicate a noticeable decline in performance, emphasizing the significant contribution of the high-resolution branch in enabling finer predictions.

% \clearpage
% \newpage

\end{document}